%% file: neurips_2026.tex
\documentclass{article}

\PassOptionsToPackage{numbers, compress}{natbib}
\usepackage[eandd,final]{neurips_2026}
\usepackage{longtable}
\usepackage{amsmath}
\usepackage{array}
\usepackage{multirow}

\usepackage[utf8]{inputenc} 
\usepackage[T1]{fontenc}    
\usepackage{hyperref}       
\usepackage{url}            
\usepackage{booktabs}       
\usepackage{amsfonts}       
\usepackage{nicefrac}       
\usepackage{microtype}      
\usepackage{xcolor}         
\usepackage{graphicx}
\usepackage{caption}
\usepackage{subcaption}
\usepackage{wrapfig}

\definecolor{gray6D}{HTML}{6D6D6D}

\title{Benchmark Health Index: A Systematic Framework for Benchmarking the Benchmarks of LLMs}

\author{%
  Longyuan Zhu\thanks{Equal contribution.} \\
  SKYEVAL \\
  Beijing, China \\
  \texttt{longyuanz@student.unimelb.edu.au} \\
  \And
  Hairan Hua\footnotemark[1] \\
  SKYEVAL \\
  Beijing, China \\
  \texttt{51295901027@stu.ecnu.edu.cn} \\
  \And
  Linlin Miao\footnotemark[1] \\
  DAMO Academy, Alibaba Group \\
  Hangzhou, Zhejiang, China \\
  \texttt{miaolinlin.mll@alibaba-inc.com} \\
  \And
  Wei Wang \\
  DAMO Academy, Alibaba Group \\
  Hangzhou, Zhejiang, China \\
  \texttt{weixiong.ww@alibaba-inc.com} \\
  \And
  Bing Zhao\thanks{Corresponding author.} \\
  DAMO Academy, Alibaba Group \\
  Zhejiang University \\
  Hangzhou, Zhejiang, China \\
  \texttt{xiongdao@alibaba-inc.com} \\
}

\begin{document}

\maketitle

\begin{abstract}
Large Language Models (LLMs) are advancing rapidly, yet the benchmarks used to measure this progress are becoming increasingly unreliable. Score inflation and selective reporting have eroded the authority of standard benchmarks, leaving the community uncertain about which evaluation results remain trustworthy. We introduce the \textbf{Benchmark Health Index (BHI)}, a pure data-driven framework for auditing evaluation sets along three complementary axes: (1) \textbf{Capability Discrimination}, measuring how sharply a benchmark separates model performance beyond noise; (2) \textbf{Anti-Saturation}, estimating remaining headroom before ceiling effects erode resolution and thus the benchmark's expected longevity; and (3) \textbf{Impact}, measuring capability-weighted benchmark adoption across the model ecosystem. By distilling \textbf{158} validated benchmarks from the technical reports of \textbf{117} representative models released between \textbf{2025} and \textbf{April 2026}, we systematically characterize the evaluation landscape. BHI is the \textbf{first framework} to quantify benchmark health at a macro level, providing a well-grounded quantitative basis for benchmark selection and supporting the dynamic filtering and continuous updating of evaluation sets. Our comprehensive analysis not only identifies benchmarks that remain highly credible but also exposes pervasive structural issues. By establishing a rigorous foundation for selection and management, BHI elevates evaluation practice from heuristic-based judgment to a quantifiable and interpretable scientific process. Our code is available online at \url{https://github.com/SKYLENAGE-AI/Benchmark_Health_Index}
\end{abstract}

\newpage
\suppressfloats[t]
\section{Introduction}

\begin{wrapfigure}{r}{0.48\linewidth}
  \centering
  \vspace{-8pt}
  \includegraphics[width=\linewidth]{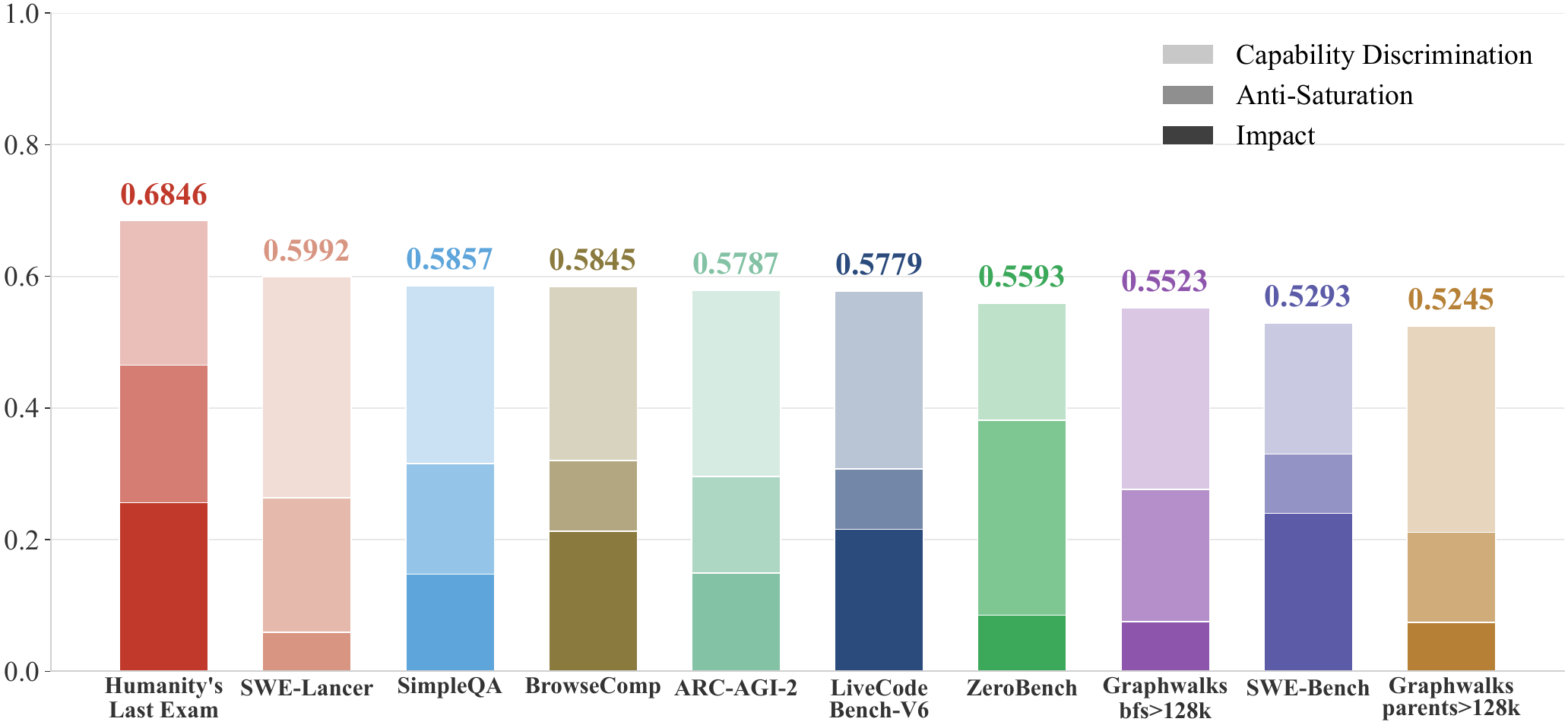}
  \caption{Top 10 Benchmarks Ranked by BHI}
  \label{fig:bhi_ranking}
  \vspace{-10pt}
\end{wrapfigure}

The progress of LLMs is inseparable from benchmarks \cite{liang2023holistic,hibench10.1145/3711896.3737378}. Benchmarks define the trajectory of scientific advancement for the research community \cite{wang2018glue, srivastava2023beyond}, provide capability references for industry, and offer essential guidance for allocating and optimizing computational resources. Ideally, a benchmark should function as a precise measurement instrument, delivering stable, reproducible, and interpretable discriminative signals across capability levels \cite{ribeiro-etal-2020-beyond}. However, the evaluation ecosystem has exhibited systematic distortions in recent years \cite{kiela-etal-2021-dynabench}. Many canonical benchmarks are approaching a compressed performance ceiling \cite{wang2019superglue}: top-tier models cluster within a narrow score band, so small gaps are often driven by statistical noise rather than substantive architectural differences. Moreover, under selective reporting and marketing-oriented practices, citation frequency no longer reliably reflects technical rigor or the long-term evaluative utility of a benchmark \cite{haimes2024benchmarkinflationrevealingllm}. These trends call for a shift from model-centric comparison to benchmark-centric auditing \cite{dehghani2021benchmarklottery}, treating benchmarks as public infrastructure with a lifecycle and quantitatively testing their validity and degradation dynamics.

Current work has laid important foundations in holistic evaluation frameworks \cite{10.1145/3722449.3722467,agentsurvey10.1145/3711896.3736570}, dataset construction quality \cite{10.1145/3711896.3737437}, and cross-benchmark consistency \cite{lunardi2025robustnessreliabilitybenchmarkbasedevaluation}. Yet most existing approaches remain limited to static specifications or local consistency analyses, lacking an actionable, quantitative, and reproducible yardstick for horizontal comparison and full-lifecycle management of benchmarks at the ecosystem scale. The community thus needs a global metric system for benchmark health: one that not only quantifies a benchmark's current discriminative capability, but also evaluates its remaining effective and ecological impact, providing a quantifiable basis for benchmark selection.

\begin{figure*}[t] 
    \centering
    \includegraphics[width=\textwidth]{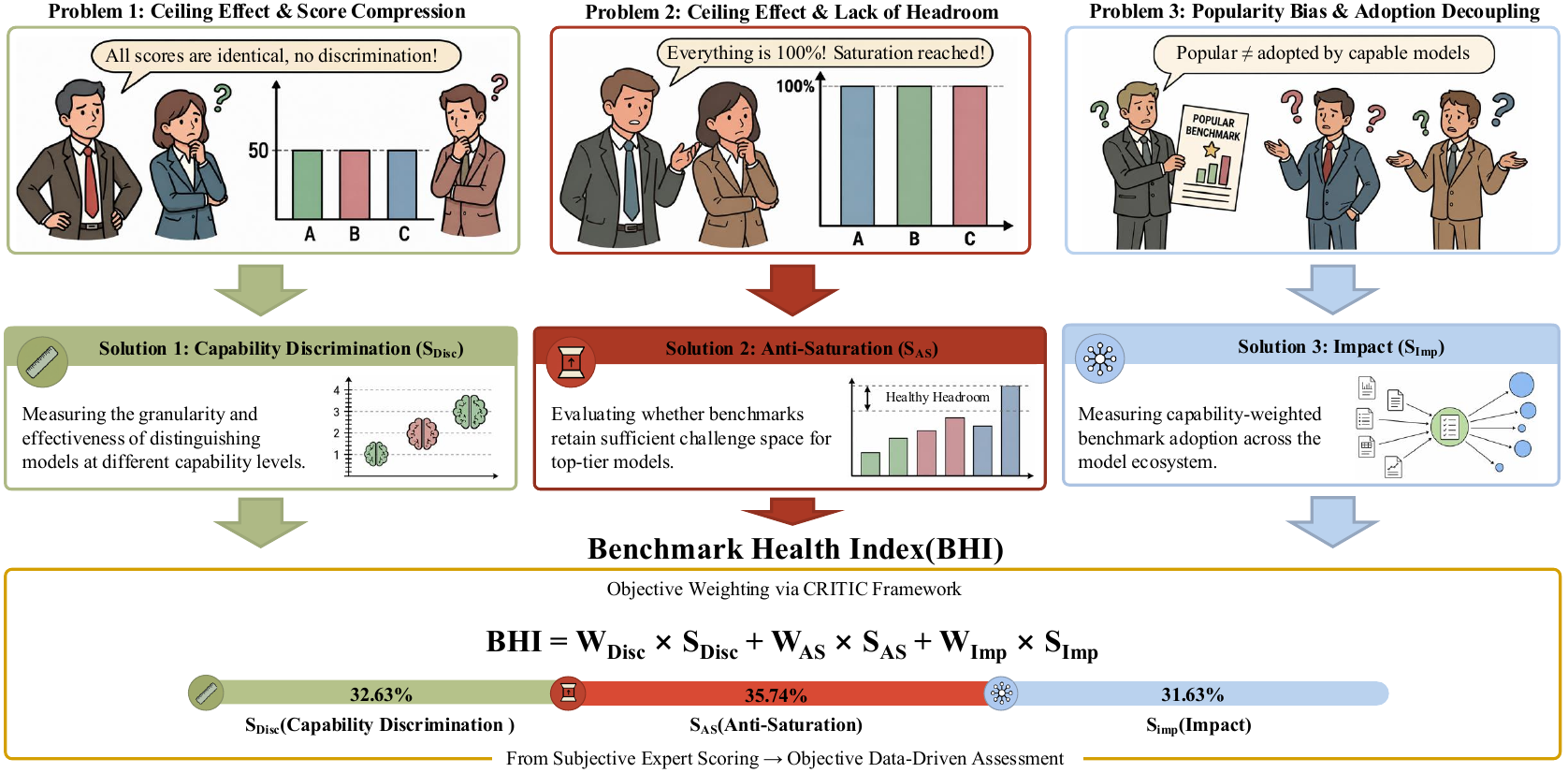} 
    \caption{Overview of BHI Framework. \textit{Top Row:} Three challenges characterize current benchmarks: weak discrimination, performance saturation, and popularity--adoption decoupling. \textit{Middle Row:} We introduce three data-driven metrics: $\boldsymbol{S_{Disc}}$ quantifies benchmark sensitivity, $\boldsymbol{S_{AS}}$ assesses challenge headroom, and $\boldsymbol{S_{Imp}}$ measures capability-weighted adoption across the model ecosystem. \textit{Bottom Row:} The final BHI score is synthesized through objective weighting.}
    \label{fig:teaser}
\end{figure*}


Formalizing benchmark health faces three technical challenges: (1) \textbf{Low Discrimination \& Score Compression}: many benchmarks can no longer clearly separate strong models because their scores are compressed near the top; (2) \textbf{Ceiling Effect \& Lack of Headroom}: as benchmark data is absorbed into training or repeatedly optimized for, the remaining challenge space shrinks, weakening long-term progress tracking; and (3) \textbf{Popularity Bias \& Adoption Decoupling}: frequent citation or reporting does not necessarily mean that a benchmark is widely adopted by high-capability models or remains technically informative. Together, these issues lead the community to overuse outdated benchmarks and overlook more valuable ones.


To address this, we introduce \textbf{Benchmark Health Index}, a data-driven auditing framework that characterizes benchmark validity and usability across three dimensions: (1) \textbf{Capability Discrimination}, which measures whether a benchmark can clearly separate stronger models from weaker ones; (2) \textbf{Anti-Saturation}, which evaluates whether a benchmark still remains challenging for current frontier models and is unlikely to be exhausted soon; and (3) \textbf{Impact}, which measures whether a benchmark is widely adopted by high-capability models in real evaluation practice. By aggregating these, BHI identifies effective benchmarks, demotes saturated tests to basic sanity checks, and designates high-difficulty anchors for frontier capability tracking.

We conducted a systematic meta-analysis of 158 benchmarks extracted from the technical reports of 117 representative LLMs released between 2025 and April 2026. Our study distinguishes effective benchmarks that retain high discriminative power from those that, despite being widely used, have become obsolete. Furthermore, we provide a systematic examination of benchmark distribution and evolutionary trends across diverse task domains.

Our contributions are summarized as follows:
\begin{itemize}
    \item We propose the BHI framework, which quantifies benchmark effectiveness and longevity along three dimensions: Capability Discrimination, Anti-Saturation, and Impact.
    \item We establish an objective, data-driven auditing protocol that implements a Leave-One-Benchmark-Out calibration strategy and automated weighting mechanisms to eliminate subjective bias and ensure methodological rigor.
    \item Through large-scale meta-analysis and diagnostic case studies over 117 models and 158 benchmarks, we show that BHI separates saturated from frontier benchmarks, explains their degradation patterns, and offers actionable guidance for benchmark governance.
\end{itemize}

\section{Related Work}
Rapid advances in LLMs have shifted evaluation from static, model-centric leaderboards toward comprehensive, multidimensional benchmark suites, such as HELM and BIG-bench \cite{liang2023holistic,srivastava2023beyond}. However, as models improve, score inflation and upper-tail compression weaken benchmarks' ability to distinguish models and track progress \cite{haimes2024benchmarkinflationrevealingllm,zhou2025lost}. Data contamination further complicates interpretation, as high scores may not reflect genuine generalization \cite{dong-etal-2024-generalization}. Evaluation therefore requires examining whether benchmarks themselves remain useful.

To mitigate benchmark degradation, existing studies explore contamination detection, scoring adjustments, and dynamic evaluation, including open-source contamination reports \cite{li-etal-2024-open-source}, weighted metrics for saturated benchmarks \cite{etzine-etal-2025-revitalizing}, Dynabench, and LatestEval \cite{kiela-etal-2021-dynabench,10.1609/aaai.v38i17.29822}. These approaches maintain evaluation utility through data reliability, model separability, and test updates, but primarily address specific failure mechanisms. A unified framework for comparing benchmark health and monitoring its evolution remains lacking. BHI addresses this gap by jointly measuring capability discrimination, anti-saturation, and capability-weighted adoption, providing a quantitative basis for benchmark selection and updating as benchmarks and models co-evolve. Appendix~\ref{appendix:expanded related work} provides a broader discussion.

\section{Methodology}
\label{sec:mathodology}
The BHI is constructed through a data-driven aggregation of three distinct evaluation axes: Capability Discrimination, Anti-Saturation, and Impact. Starting from normalized model scores, we combine pairwise score differences and distributional spread to measure discrimination. We then estimate model capabilities using a Leave-One-Benchmark-Out (LOBO) strategy and use these estimates to weight remaining headroom and benchmark adoption. SDM combines the two discrimination indicators, while CRITIC aggregates the three axes into the final BHI. The following subsections define each component, with the final aggregation presented in Section~\ref{sec:main_results}.

\subsection{Capability Discrimination ($S_{Disc}$)}
$S_{Disc}$ quantifies a benchmark $b \in B$'s resolution in distinguishing latent model performance through \textbf{micro-resolution} and \textbf{macro-bandwidth}. All model scores are pre-normalized to a $[0, 100]$ scale for consistency.

\subsubsection{Effective Differentiation Ratio (EDR)} 
$EDR$ measures micro-resolution as the proportion of all possible pairwise combinations of models whose performance gap exceeds a noise-filtering threshold. For $N$ models, it is defined as:
\begin{equation}
\label{equ:EDR}
EDR(b) = \frac{\sum_{1\le i < j \le N} \mathbb{I}(|\text{Score}_i - \text{Score}_j| > \delta)}{N(N-1)/2}
\end{equation}
where $\delta = 0.02 \cdot (\max(\text{Score}) - \min(\text{Score}))$ is an adaptive threshold representing 2\% of the observed score range, filtering small pairwise differences relative to the benchmark's performance spread. To assess sensitivity to this choice, we vary the threshold from 0.5\% to 5.0\% of the observed range and recompute the full BHI pipeline. Compared with the default ranking, Spearman correlations range from 0.9582 to 1.0000, with 9--10 benchmarks retained in the top 10. Correlations exceed 0.995 for thresholds between 1.5\% and 2.5\%, indicating stable rankings around the default setting. Appendix~\ref{app:edr_threshold_sensitivity} provides the complete results and analysis.


\begin{figure*}[t]
    \centering
    \includegraphics[width=\textwidth]{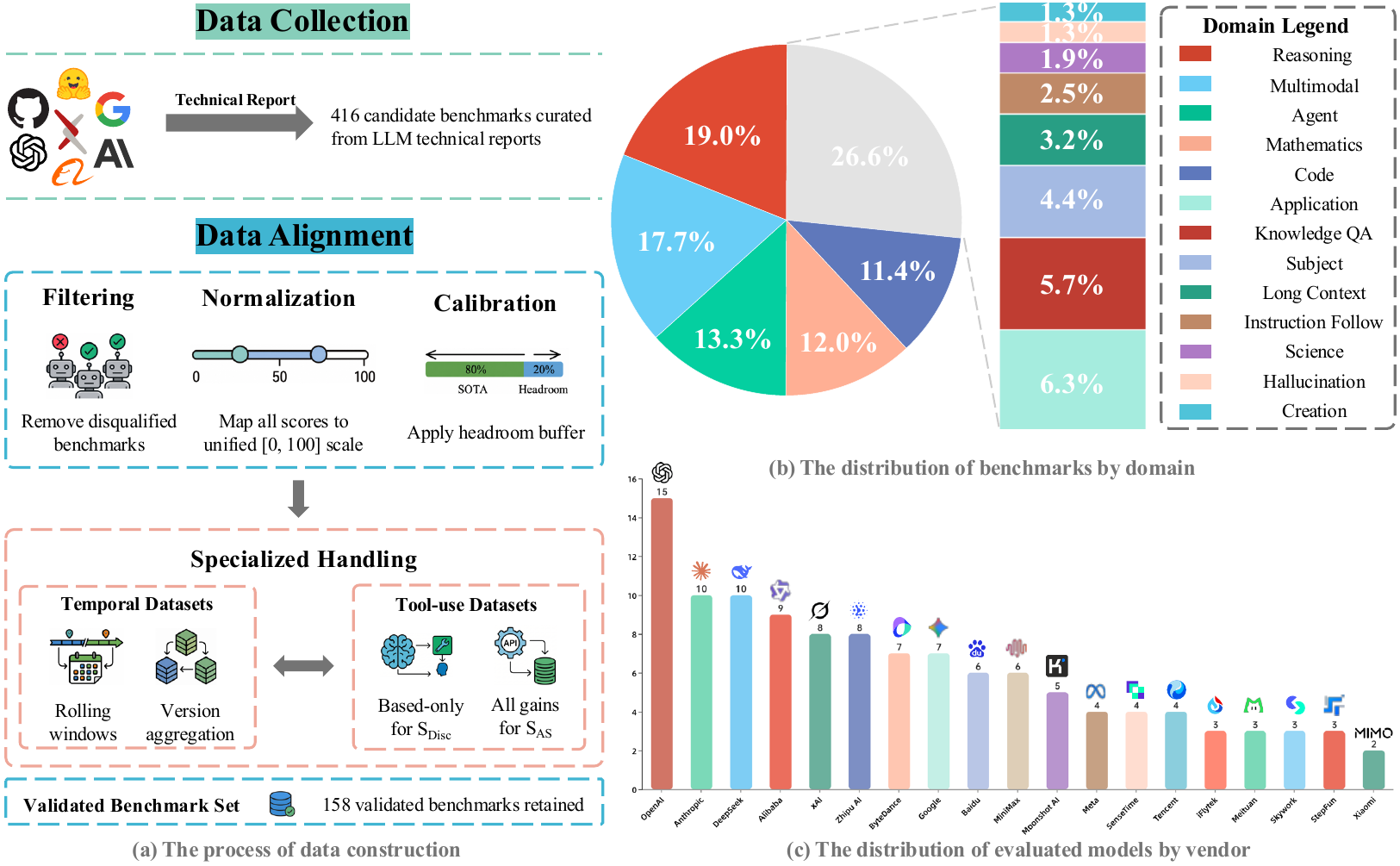}
    \caption{The BHI Data Architecture. (a) The systematic process of data construction. (b) The taxonomic distribution of the validated benchmark set across 13 functional domains. (c) The distribution of evaluated models across diverse global AI vendors.}
    \label{fig:data_pipeline}
\end{figure*}

\subsubsection{Robust Coefficient of Variation (RCV)}

$RCV$ characterizes macro-bandwidth by measuring the effective performance spread using the middle 80\% of the score distribution for the $N$ models, thereby mitigating the influence of extreme outliers:
\begin{equation}
RCV(b) = \frac{P_{90}(\text{Score}) - P_{10}(\text{Score})}{100}
\end{equation}

To assess sensitivity to this choice, we test six percentile intervals from P0--P100 to P25--P75 and recompute the BHI rankings. Compared with the default P10--P90 setting, Spearman correlations remain at least 0.9922, with at least 9 of the top 10 benchmarks retained and median absolute rank shifts of at most three positions. Complete results are provided in Appendix~\ref{app:rcv_sensitivity}.

\subsubsection{Normalization} 
Prior to final fusion, we apply min-max normalization to both EDR and RCV to map them onto a common interval $[0, 1]$. To ensure the final score accurately reflects the informative contribution of each metric, we aggregate these indicators using Standard Deviation-based Weighting (SDM). This approach amplifies relative contrast while mitigating statistical variance shrinkage in benchmarks with sparse model coverage. The final score for benchmark $b$ is computed as:
\begin{equation}
S_{Disc}(b) = \sum_{k \in \{EDR, RCV\}} w_k \cdot \text{Norm}(k)
\end{equation}
Detailed mathematical derivations for the SDM weighting and the normalization process are provided in Appendix~\ref{appendix:sdm_derivation}.

\subsection{Model Capability Calibration}
\label{sec:ModelCapabilityCalibrationwithLOBOStrategy}
To provide a fair reference for Anti-Saturation (Section \ref{sec:antisaturation}) and Impact (Section \ref{sec:impact}), we establish a system to calibrate model capabilities. To prevent self-referential bias caused by evaluating a benchmark using model strengths derived from the same data pool, we implement a Leave-One-Benchmark-Out (LOBO) strategy where capability scores $\theta_i^{(-b)}$ for any target benchmark $b \in B$ are calculated strictly using data from the remaining $B \setminus \{b\}$ benchmarks, thereby ensuring methodological independence through out-of-sample evaluation.

\subsubsection{LOBO-adjusted Win Rate.} 
We leverage a pairwise battle mechanism to ensure a robust and unified standardization of scores across the diverse array of metrics. For model $i$ on benchmark $b$, its LOBO-adjusted Win Rate ($W_{i}^{(-b)}$) represents the average probability of defeating all opponents across the set of benchmarks excluding $b$:
\begin{equation}
W_{i}^{(-b)} = \frac{\sum_{b' \in B \setminus \{b\}} (\text{Win}_{i,b'} + 0.5 \cdot \text{Tie}_{i,b'})}{\sum_{b' \in B \setminus \{b\}} (N_{b'} - 1)}
\end{equation}
where $\text{Win}_{i,b'}$ and $\text{Tie}_{i,b'}$ denote the number of wins and ties of model $i$ on benchmark $b'$, and $N_{b'}$ is the total number of models evaluated on $b'$. This win rate ensures consistency across heterogeneous evaluation scales.

\subsubsection{The Fourth-root Log-Balance Model.} 
We integrate the win rate with the model's participation density to ensure that capability scores rigorously reflect both model performance and statistical reliability. For benchmark $b$, the calibrated capability of model $i \in N$ is defined as:
\begin{equation}
\theta_i^{(-b)} = W_{i}^{(-b)} \cdot \left[ \frac{\ln(1 + n_{i}^{(-b)})}{\ln(1 + N_{B})} \right]^{1/4}
\end{equation}
where $n_{i}^{(-b)}$ is the number of benchmarks model $i$ has participated in within the LOBO reference pool $B \setminus \{b\}$, and $N_{B}$ is the total number of benchmarks in that reference pool (i.e., $|B| - 1$). The constant $1$ is added to logarithmic terms for numerical stability.


\subsection{Anti-Saturation ($S_{AS}$)}
\label{sec:antisaturation}
Anti-saturation characterizes the structural durability of a benchmark $b \in B$ by measuring its remaining performance headroom under the pressure of evolving model capabilities. This dimension is decomposed into two components: \textbf{Static Weighted Resistance} and \textbf{Dynamic Saturation Projection}.

\subsubsection{Static Weighted Resistance ($S_{Sta}$)}
$S_{Sta}$ estimates a benchmark's difficulty by capability-weighting the score distribution. We utilize $\theta_i^{(-b)}$ (explained in Section \ref{sec:ModelCapabilityCalibrationwithLOBOStrategy}) to evaluate benchmark difficulty relative to the strength of its participants. The static resistance is defined as:
\begin{equation}
S_{Sta}(b) = 1 - \frac{\sum_{i \in N} \text{Score}_{i,b} \cdot \theta_i^{(-b)}}{\sum_{i \in N} \theta_i^{(-b)}}
\end{equation}
where $\text{Score}_{i,b} \in [0, 1]$ denotes the normalized performance. This weighted approach applies differential penalties to the remaining headroom, whereby scores from low-capability models induce a more pronounced reduction in $S_{Sta}$ while those from frontier models are comparatively preserved. Normalizing by $\sum \theta_i^{(-b)}$ further eliminates the influence of participant volume, ensuring that $S_{Sta}$ remains invariant to the total number of evaluated models.

\subsubsection{Dynamic Saturation Projection ($S_{{Dyn}}$)}
\begin{wrapfigure}{r}{0.52\linewidth}
    \centering
    \vspace{-8pt}
    \includegraphics[width=\linewidth]{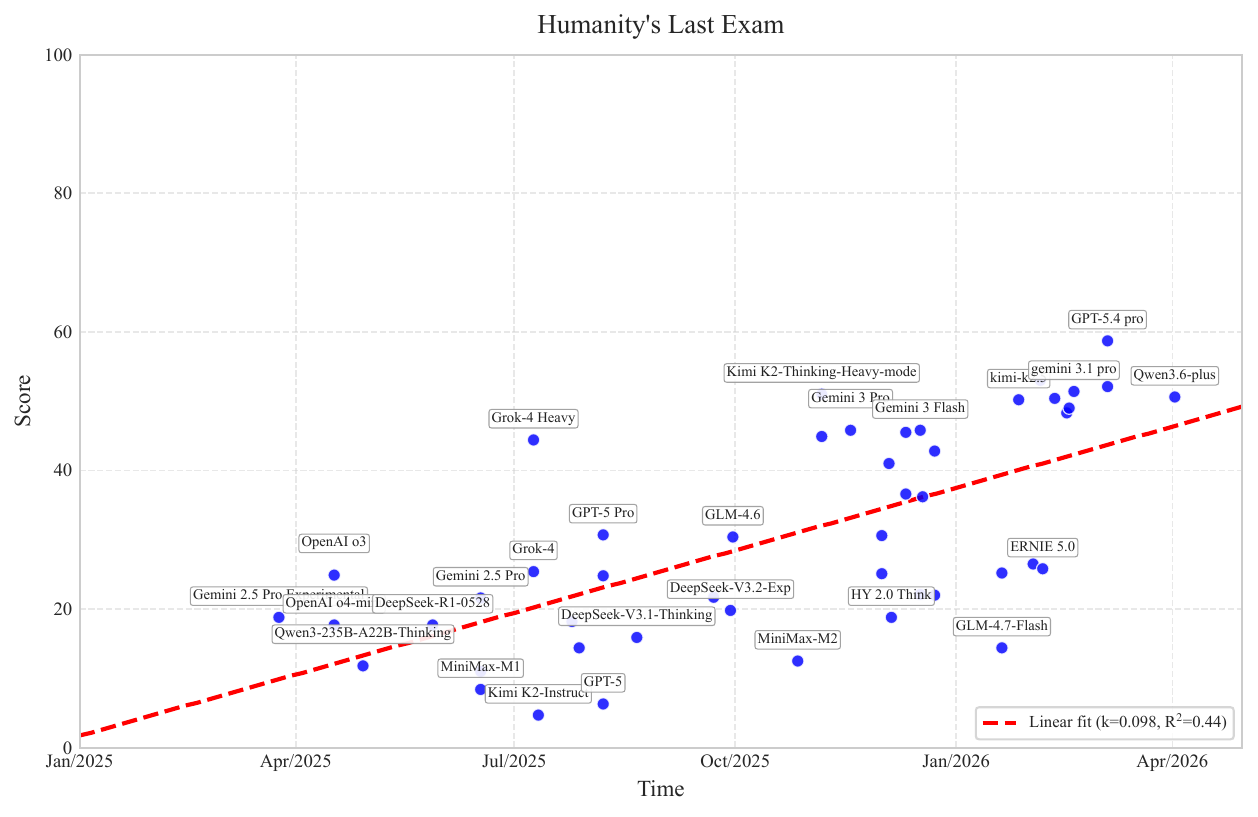}
    \caption{Performance evolution on Humanity's Last Exam. The dashed line represents the fitted linear trend, whose slope ($k$) indicates the rate of score improvement over time.}
    \label{fig:single_col_trend}
    \vspace{-10pt}
\end{wrapfigure}

$S_{{Dyn}}$ measures the short-term velocity of score inflation by projecting near-future performance under a time-series trend model.
We fit a linear trend $y = k \cdot t + d$ using Ordinary Least Squares, as shown in Figure \ref{fig:single_col_trend}, where the slope $k$ represents the \textbf{conquest rate} (score increment per day). Using a 30-day projection window, the dynamic saturation is defined as:
\begin{equation}
S_{Dyn}(b) = \max(0, 1 - (\overline{\text{Score}}_{b} + 30 \cdot k))
\end{equation}
where $\overline{\text{Score}}_{b}$ is the current mean score of benchmark $b$. If $k < 0$, typically indicating the recent entry of lower-performing models, we set $S_{Dyn} = S_{Sta}$ to maintain logical consistency. 


The final Anti-Saturation score for benchmark $b$ is computed as:
\begin{equation}
S_{AS}(b) = 0.8 \cdot S_{Sta} + 0.2 \cdot S_{Dyn}
\end{equation}
We assign a lower weight ($20\%$) to $S_{Dyn}$ due to the inherent uncertainty of short-term projections, while the static component provides a stable observation of the current saturation status.

\subsection{Impact ($S_{Imp}$)}
\label{sec:impact}

$S_{Imp}$ measures the ecological influence of a benchmark $b \in B$ through its adoption by capable models. 
Rather than relying on raw popularity signals such as repository stars, downloads, or historical citation momentum, we define impact as capability-weighted evaluation coverage within the model ecosystem.

\subsubsection{Capability-weighted Adoption}

Let $\mathcal{M}_{\text{bench}} \subseteq \mathcal{M}_{\text{all}}$ denote the set of models that have reported valid evaluation results on benchmark $b$, where $\mathcal{M}_{\text{all}}$ is the full set of models in the dataset. 
A model is considered to participate in $b$ if it reports a valid score, regardless of the score magnitude.

To avoid circularity, we use the LOBO-calibrated capability score $\theta_i^{(-b)}$ defined in Section~3.2, which is computed by excluding benchmark $b$ from the capability estimation process. 
The raw capability coverage of benchmark $b$ is defined as:
\begin{equation}
S_{Imp}^{raw}(b) =
\frac{\sum_{i \in \mathcal{M}_{\text{bench}}} \theta_i^{(-b)}}
{\sum_{j \in \mathcal{M}_{\text{all}}} \theta_j^{(-b)}} .
\end{equation}

We then apply a square-root transformation:
\begin{equation}
S_{Imp}(b) = \sqrt{S_{Imp}^{raw}(b)} .
\end{equation}

The square-root mapping compresses the upper range of the coverage ratio, preventing highly reported benchmarks from dominating solely due to participation volume, while still preserving the advantage of benchmarks evaluated by high-capability models.

\subsubsection{Interpretation}

This formulation captures impact as \emph{capability-weighted adoption}. 
A benchmark receives a high $S_{Imp}$ when a substantial portion of the ecosystem's calibrated model capability is represented among its evaluated participants. 
Because the capability scores are computed using LOBO calibration, the benchmark being assessed does not contribute to the capability estimates used to measure its own impact.

By construction, $S_{Imp}$ is bounded within $[0,1]$ and does not depend on benchmark release dates, model release dates, community metadata, or heuristic temporal decay. 
This design makes the impact score less sensitive to raw popularity accumulation and better aligned with the technical adoption of benchmarks by capable models.

\begin{table*}[t]
\centering
\caption{BHI Metrics and Sub-metric Decomposition for Representative Benchmarks}
\label{tab:bhi_full_decomposition_case_study}
\resizebox{\textwidth}{!}{%
\small
\begin{tabular}{llcccc}
\toprule
Rank & Benchmark & \textbf{$S_{\mathrm{Disc}}$} ($RCV$, $EDR$) & \textbf{$S_{\mathrm{AS}}$} ($S_{\mathrm{Sta}}$, $S_{\mathrm{Dyn}}$) & \textbf{$S_{\mathrm{Imp}}$} ($N_{\mathrm{Usage}}$) & \textbf{BHI} \\
\midrule
1  & Humanity's Last Exam\cite{phan2025humanitysexam} & \textbf{0.6608} (0.2954, 0.9131) & \textbf{0.6299} (0.6211, 0.6653) & \textbf{0.7710} (56) & \textbf{0.6846} \\
3  & SimpleQA\cite{wei2024measuringshortformfactualitylarge}             & \textbf{0.8130} (0.4599, 0.9347) & \textbf{0.5043} (0.5011, 0.5169) & \textbf{0.4435} (27) & \textbf{0.5857} \\
4  & BrowseComp\cite{wei2025browsecompsimplechallengingbenchmark}           & \textbf{0.8042} (0.4490, 0.9365) & \textbf{0.3281} (0.3200, 0.3607) & \textbf{0.6475} (36) & \textbf{0.5845} \\
7  & ZeroBench\cite{roberts2025zerobenchimpossiblevisualbenchmark}            & \textbf{0.5205} (0.1925, 0.8000) & \textbf{0.8687} (0.8594, 0.9059) & \textbf{0.2498} (6)  & \textbf{0.5593} \\
16 & AIME 2025\cite{aime25}            & \textbf{0.6923} (0.3445, 0.8889) & \textbf{0.1154} (0.1122, 0.1284) & \textbf{0.7194} (61) & \textbf{0.4947} \\
20 & Tau2-bench-Telecom\cite{barres2025tau2}   & \textbf{0.9005} (0.6419, 0.7800) & \textbf{0.1108} (0.1058, 0.1309) & \textbf{0.5016} (25) & \textbf{0.4921} \\
43 & GDPval-AA Elo\cite{patwardhan2025gdpvalevaluatingaimodel}        & \textbf{0.5962} (0.2219, 0.9111) & \textbf{0.3362} (0.3479, 0.2897) & \textbf{0.3409} (9)  & \textbf{0.4225} \\
57 & MMLU-Pro\cite{10.5555/3737916.3740934}             & \textbf{0.4572} (0.1443, 0.7520) & \textbf{0.1727} (0.1704, 0.1822) & \textbf{0.5724} (47) & \textbf{0.3919} \\
113& HumanEval\cite{HumanEval+10.5555/3666122.3667065}            & \textbf{0.5520} (0.1243, 1.0000) & \textbf{0.1669} (0.1710, 0.1508) & \textbf{0.1232} (4)  & \textbf{0.2787} \\
140& GSM8K\cite{cobbe2021trainingverifierssolvemath}                & \textbf{0.3920} (0.1368, 0.6222) & \textbf{0.0865} (0.0835, 0.0989) & \textbf{0.1931} (10) & \textbf{0.2199} \\
\bottomrule
\end{tabular}%
}
\end{table*}

\section{Data Pipeline}
\label{data_pipline}
As illustrated in Figure~\ref{fig:data_pipeline}, we implement a multi-stage pipeline to distill a high-fidelity benchmark suite from an initial corpus of 416 candidate evaluation sets, systematically aggregated from the official technical reports of 117 mainstream models released between 2025 and April 2026. The workflow comprises data collection, alignment, and validation.

The alignment stage (Figure~\ref{fig:data_pipeline}a) excludes benchmarks evaluated by fewer than three models to support dispersion-based estimation. Scores are normalized to a $[0, 100]$ range, with lower-is-better metrics transformed to a higher-is-better orientation. For open-ended metrics lacking a theoretical maximum, we normalize scores against 1.25 times the observed maximum, mapping the current best performance to 80 and reserving 20 points of headroom. For temporal benchmarks such as LiveCodeBench, fragmented snapshots are aggregated into major version clusters under the assumption of difficulty invariance. For tool-augmented evaluations, Capability Discrimination uses base-ability scores, whereas Anti-Saturation incorporates performance gains from agentic and retrieval-based optimizations to assess the practical exhaustion of the task environment.

After alignment and filtering, we retain 158 validated benchmarks spanning diverse domains and vendors (Figure~\ref{fig:data_pipeline}b--c), including reasoning, multimodal understanding, mathematics, and code generation, with coverage of both open- and closed-source models. Appendix~\ref{append:Expanded Data Pipeline} provides the complete filtering, normalization, and calibration procedures. The complete metadata is publicly available at our anonymous repository.

\section{Results and Validation}
\label{results_and_validation}
\subsection{Main BHI Results}
\label{sec:main_results}

The final BHI is synthesized from the three primary dimensions introduced in Section~\ref{sec:mathodology}: $S_{Disc}$, $S_{AS}$, and $S_{Imp}$. 
In this process, the $S_{AS}$ and $S_{Imp}$ axes are adjusted by the Model Capability Calibration component---derived from the LOBO-adjusted win rate---to transform heterogeneous benchmark-level signals into a unified health score.

We utilize the CRITIC objective weighting framework, as detailed in Appendix~\ref{appendix:critic_derivation}, for the final index aggregation to maintain a consistent data-driven formulation. 
This method determines the importance of each dimension by evaluating its internal information volume, measured by contrast intensity, and its degree of independence, measured by conflicting character, within the global distribution of 158 benchmarks. 
Under this framework, the optimized weighting distribution is 32.63\% for $S_{Disc}$, 35.74\% for $S_{AS}$, and 31.63\% for $S_{Imp}$, resulting in the final formulation:
\begin{equation}
\label{bhi_equation}
BHI(b) = w_{Disc} \cdot S_{Disc}(b) + w_{AS} \cdot S_{AS}(b) + w_{Imp} \cdot S_{Imp}(b)
\end{equation}

Representative scores and rankings are summarized in Table~\ref{tab:bhi_full_decomposition_case_study}, while the comprehensive rankings for the complete set of 158 benchmarks are provided in Table~\ref{tab:appendix-bhi_full_decomposition_case_study}. 
Overall, the resulting ranking reflects a balance between technical resolution, saturation resistance, and capability-weighted adoption, rather than optimizing for any single benchmark property.

\subsection{Component Contributions: Ablation Study}

To assess the contribution of each component in the BHI formulation, we conduct an ablation study by removing one indicator at a time and recomputing the final ranking. 
The ranking differences are measured relative to the full model using average and maximum absolute rank shifts.


\begin{wraptable}{r}{0.52\textwidth}
\vspace{-1.0em}
\centering
\caption{Ablation results showing ranking shifts after removing each component.}
\label{tab:ablation}
\begin{tabular*}{\linewidth}{@{\extracolsep{\fill}}lcc@{}}
\toprule
\textbf{Removed Component} & \textbf{Avg. Shift} & \textbf{Max Shift} \\
\midrule
$S_{Disc}$ & 15.53 & 49 \\
$S_{AS}$   & 16.18 & 82 \\
$S_{Imp}$  & 11.34 & 66 \\
\bottomrule
\end{tabular*}
\vspace{-0.8em}
\end{wraptable}

As shown in Table~\ref{tab:ablation}, removing any single component leads to substantial ranking changes. 
Excluding $S_{AS}$ results in the largest maximum shift and may favor near-ceiling benchmarks with strong discrimination or adoption. 
Removing $S_{Disc}$ produces a comparable average shift and may favor benchmarks with limited model separation but strong headroom or adoption. 
Although removing $S_{Imp}$ yields a smaller average shift, it still changes individual benchmark rankings and may favor technically strong benchmarks with limited capability-weighted adoption.

These results suggest that the three components contribute complementary diagnostic information to the BHI ranking.

\subsection{Component Complementarity}

\begin{figure}[h]
\centering
\includegraphics[width=0.95\linewidth]{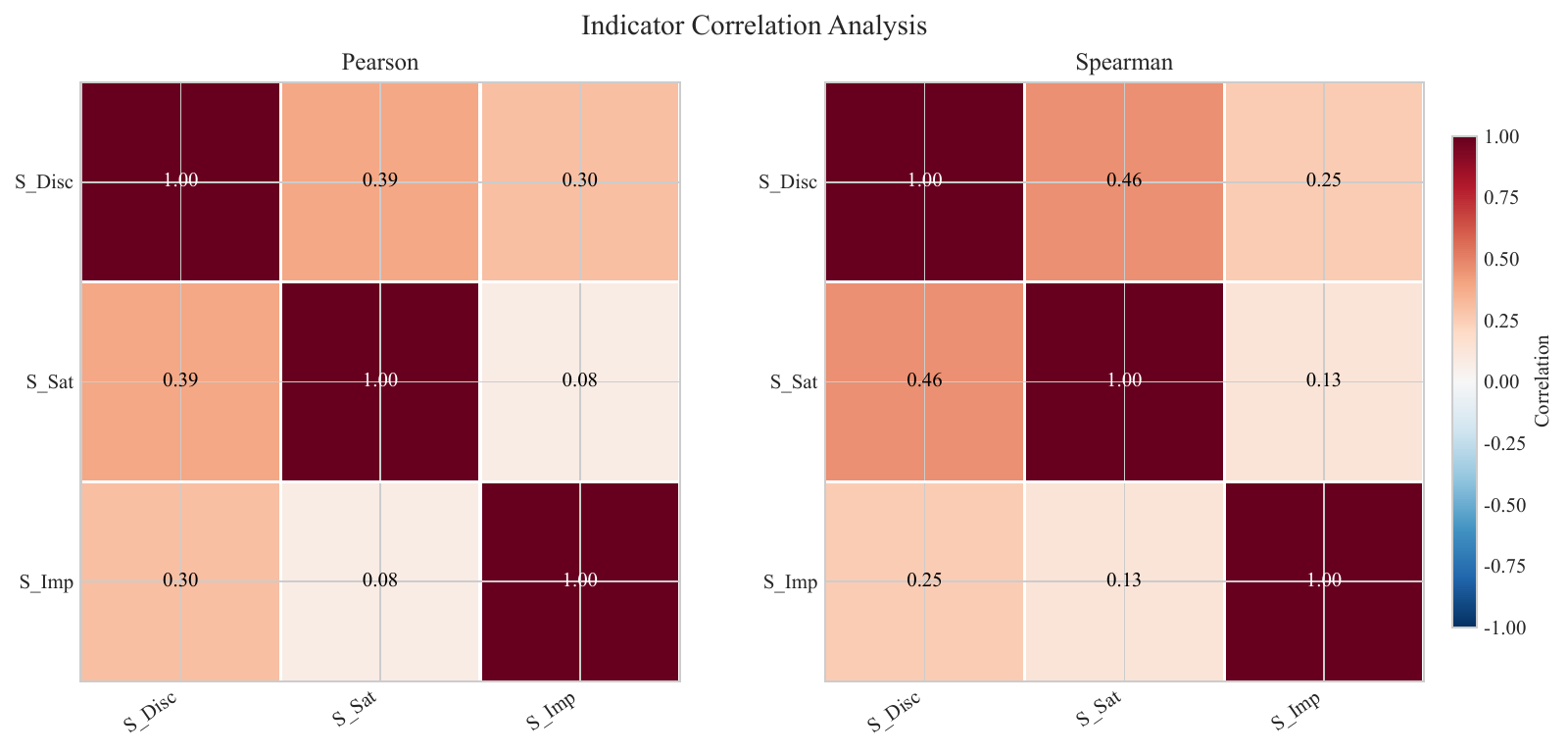}
\caption{Indicator correlation analysis using Pearson (left) and Spearman (right) coefficients.}
\label{fig:correlation}
\end{figure}

We further analyze the relationships between the three indicators using both Pearson and Spearman correlation coefficients. 
The results are visualized in Figure~\ref{fig:correlation}.

The correlation analysis suggests that the indicators are complementary rather than redundant. 
The strongest association occurs between $S_{Disc}$ and $S_{AS}$ (Spearman $\rho=0.46$), which is expected since highly saturated benchmarks tend to exhibit reduced discriminative resolution. 
A uniform upward shift in scores can leave their spread unchanged while reducing remaining headroom. Conversely, scores clustered near the floor may indicate substantial headroom but limited discrimination.
However, all pairwise correlations remain below 0.5, and $S_{Imp}$ shows consistently weak correlations with both technical indicators.

This indicates that $S_{Imp}$ captures an adoption-oriented signal distinct from benchmark difficulty and discrimination, while $S_{Disc}$ and $S_{AS}$ provide related but non-identical views of benchmark quality. 
Overall, the results support the complementary roles of the three components in characterizing benchmark health.

\paragraph{Robustness and Sensitivity.}
We evaluate the robustness of the proposed framework under multiple perturbations, including score noise, model dropout, parameter variation, and aggregation schemes. 
Under Gaussian noise ($\sigma \in [0.01, 0.2]$), the resulting rankings remain stable, with Spearman correlations remaining above 0.9 under moderate perturbation and degrading only gradually as the noise level increases; detailed settings and results are provided in Appendix~\ref{app:robustness_noise_dropout}. 
Similarly, random model dropout from $5\%$ to $60\%$ produces only gradual ranking degradation, indicating resilience to incomplete evaluation coverage; the full dropout results are also reported in Appendix~\ref{app:robustness_noise_dropout}.

We further analyze sensitivity to key design choices in the anti-saturation component. 
Varying the static/dynamic weighting in $S_{\mathrm{AS}}$ and the temporal projection window used in $S_{\mathrm{Dyn}}$ produces negligible changes in the final ranking, with Spearman correlations above 0.99 across all tested configurations, as detailed in Appendix~\ref{app:antisaturation_sensitivity}. 
In addition, the EDR threshold used in $S_{\mathrm{Disc}}$ is varied across $\delta \in [0.5\%, 5.0\%]$, and the resulting rankings remain highly consistent with the default setting; this threshold sensitivity analysis is provided in Appendix~\ref{app:threshold_sensitivity}. 
We also compare the CRITIC objective weighting scheme with uniform weighting and observe that CRITIC yields consistently more stable rankings under both noise injection and model dropout, supporting its use as a data-driven aggregation strategy; the comparative results are included in Appendix~\ref{app:robustness_noise_dropout}.

To examine potential bias from data availability, we analyze the relationship between benchmark coverage and BHI scores. 
Although coverage is strongly correlated with $S_{\mathrm{Imp}}$ by construction, its correlation with $S_{\mathrm{Disc}}$ and $S_{\mathrm{AS}}$ remains weak, and the overall correlation with BHI is moderate, indicating that the final ranking is not dominated by benchmark popularity or evaluation frequency; detailed coverage-bias results are reported in Appendix~\ref{app:coverage_bias}. 
Overall, these results demonstrate that the BHI framework is robust to score noise, model sparsity, parameter choices, and coverage variation.

\section{External Validation and Case Study}

To examine whether BHI is consistent with external evidence, we conduct a diagnostic case study using two benchmark cohorts defined independently of the BHI computation. The known-saturated cohort includes benchmarks widely regarded as losing frontier-level diagnostic value due to ceiling effects, score compression, leakage concerns, or replacement by stronger successors, while the active frontier cohort includes benchmarks that remain in recent frontier-model reports for reasoning, browsing, coding, and agentic evaluation. These cohort labels are used only for validation and are never used to compute BHI scores.

\begin{wrapfigure}{r}{0.48\linewidth}
    \centering
    \vspace{-0.8em}
    \includegraphics[width=\linewidth]{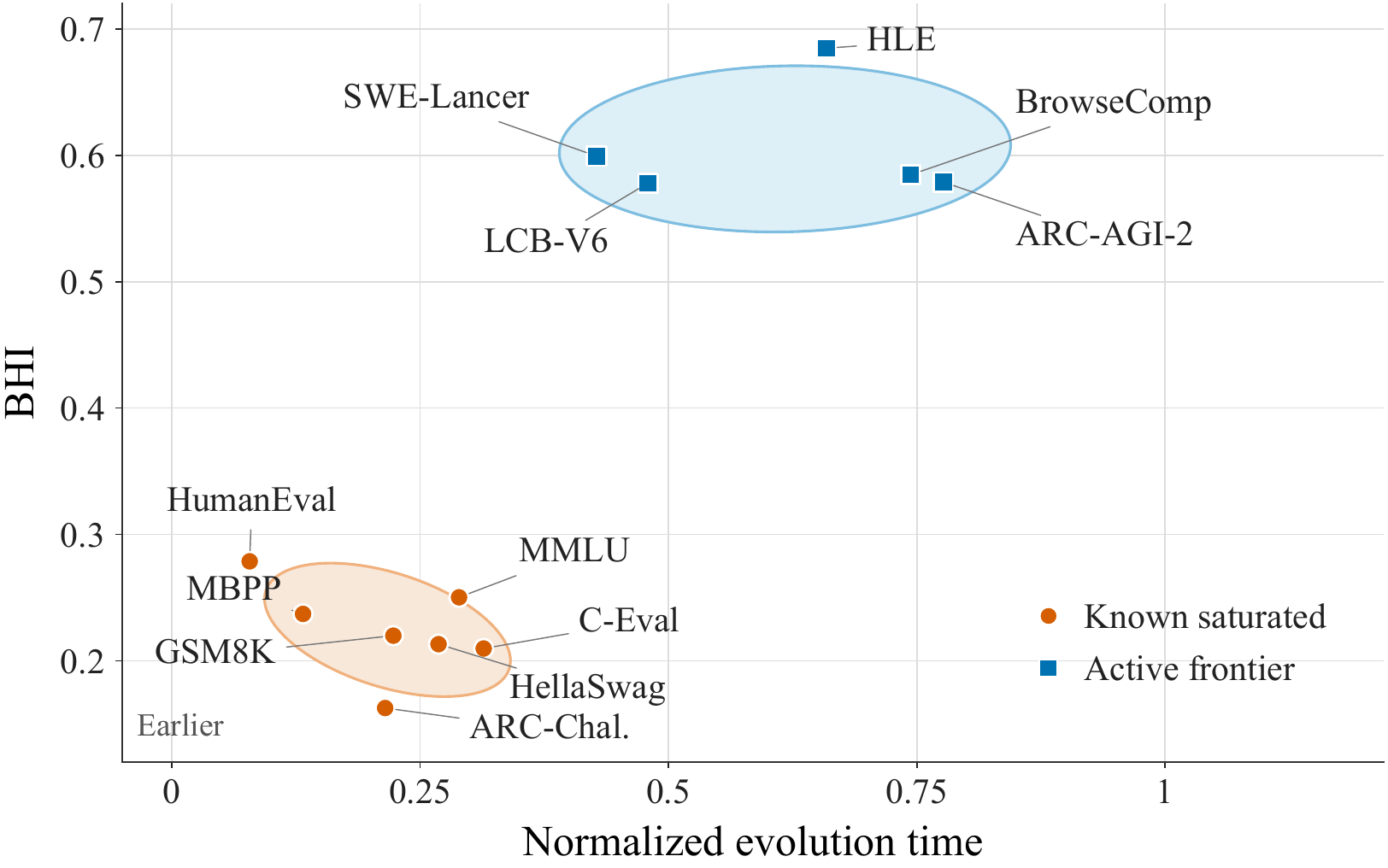}
    \caption{Temporal adoption patterns of known-saturated and active frontier benchmarks.}
    \label{fig:case_temporal_adoption}
    \vspace{-1.2em}
\end{wrapfigure}

Figure~\ref{fig:case_temporal_adoption} provides a temporal check: known-saturated benchmarks show reduced presence in recent frontier reports, whereas active frontier benchmarks remain in contemporary evaluation practice. Table~\ref{tab:validity} further shows a clear separation between the two groups, with known-saturated benchmarks averaging $0.225 \pm 0.036$ and active frontier benchmarks averaging $0.605 \pm 0.045$. This supports directional consistency with external practice, though it should be interpreted as face validity rather than calibration against an absolute ground-truth health score.

The component decomposition further explains why benchmarks receive low BHI scores. HellaSwag and ARC-Challenge are mainly penalized by severe ceiling effects, reflected in extremely low $S_{AS}$. HumanEval and MBPP retain some discriminative signal but have low anti-saturation scores, suggesting that function-level coding tasks have largely been absorbed by contemporary training and evaluation practice. C-Eval degrades across all three axes, consistent with its reduced reliability as a modern subject-knowledge benchmark. Importantly, GSM8K and MMLU still retain non-trivial $S_{Imp}$, indicating that their low BHI scores are not merely caused by low visibility or sparse reporting, but are mainly saturation-driven. This decomposition shows that BHI not only ranks benchmarks, but also distinguishes failure modes including saturation-driven obsolescence, weak adoption, and reduced discrimination.

\begin{table}[h]
\centering
\caption{BHI Distribution Across Known-Saturated and Active Frontier Benchmark Cohorts}
\label{tab:validity}
\begin{tabular}{llccccc}
\toprule
\textbf{Category} & \textbf{Benchmark} & $S_\text{Disc}$ & $S_\text{AS}$ & $S_\text{Imp}$ & \textbf{BHI} & \textbf{Rank} \\
\midrule
\multirow{7}{*}{Known Saturated}
 & HumanEval\cite{HumanEval+10.5555/3666122.3667065}          & 0.5520 & 0.1669 & 0.1232 & 0.2787 & 113 \\
 & MMLU\cite{hendrycks2021measuring}               & 0.3215 & 0.1193 & 0.3248 & 0.2502 & 126 \\
 & MBPP\cite{mbppplus10.1145/3680410}               & 0.4225 & 0.1744 & 0.1168 & 0.2371 & 129 \\
 & GSM8K\cite{cobbe2021trainingverifierssolvemath}              & 0.3920 & 0.0865 & 0.1931 & 0.2199 & 140 \\
 & HellaSwag\cite{zellers-etal-2019-hellaswag}          & 0.4395 & 0.0835 & 0.1263 & 0.2132 & 142 \\
 & C-Eval\cite{huang2023ceval}             & 0.2674 & 0.0935 & 0.2816 & 0.2098 & 143 \\
 & ARC-Challenge\cite{leolm2023arc_de}      & 0.3395 & 0.0430 & 0.1155 & 0.1627 & 155 \\
\midrule
\multirow{5}{*}{Active Frontier}
 & Humanity's Last Exam\cite{phan2025humanitysexam}  & 0.6608 & 0.6299 & 0.7710 & 0.6846 & 1 \\
 & SWE-Lancer\cite{miserendino2025swelancerfrontierllmsearn}            & 0.9979 & 0.6102 & 0.1757 & 0.5992 & 2 \\
 & BrowseComp\cite{wei2025browsecompsimplechallengingbenchmark}            & 0.8042 & 0.3281 & 0.6475 & 0.5845 & 4 \\
 & ARC-AGI-2\cite{chollet2026arcagi2newchallengefrontier}             & 0.8531 & 0.4411 & 0.4513 & 0.5787 & 5 \\
 & LiveCodeBench-V6\cite{jain2024livecodebench}      & 0.8266 & 0.2786 & 0.6593 & 0.5779 & 6 \\
\bottomrule
\end{tabular}
\end{table}

Overall, these case studies support BHI as a data-driven benchmark auditing framework whose outputs align with external evaluation practice while remaining interpretable at the component level. More detailed domain-specific analyses are provided in Appendix~\ref{append:doman-specific-analyisis}.

\section{Conclusion}

This paper introduces the \textbf{Benchmark Health Index (BHI)}, a data-driven framework for auditing LLM benchmarks at the ecosystem level. Unlike conventional model-centric leaderboards, BHI treats benchmarks themselves as objects of evaluation and quantifies their health through discrimination, anti-saturation, and capability-weighted adoption. Based on large-scale public technical reports, BHI produces interpretable benchmark rankings and reveals current failure modes in the evaluation ecosystem. Validation further shows that its components are complementary and stable. Overall, BHI provides a reproducible quantitative basis for benchmark selection, lifecycle monitoring, and systematic benchmark governance.

\section{Acknowledgements}
This work was supported by SKYEVAL and Alibaba DAMO Academy.


\newpage
\bibliographystyle{myunsrtnat}
\bibliography{sample-base}

\newpage
\appendix

\section{Limitations}
\label{app:limitations}

BHI is, to the best of our knowledge, the first benchmark-level auditing framework that systematically quantifies benchmark health across discrimination, anti-saturation, and ecosystem impact, rather than an absolute measure of model capability or benchmark quality. Its conclusions should therefore be interpreted as relative diagnostic signals within the collected evaluation ecosystem. The framework is most informative when a benchmark has sufficient public reporting coverage and when the reported scores are comparable after normalization. Benchmarks with limited adoption, highly specialized evaluation protocols, or rapidly changing versions may receive less stable estimates, even though the LOBO calibration in Section~\ref{sec:ModelCapabilityCalibrationwithLOBOStrategy} and the coverage analysis in Appendix~\ref{app:coverage_bias} are designed to reduce this effect.

A second limitation comes from the heterogeneity of public technical reports. Although the data pipeline in Section~\ref{sec:mathodology} and Appendix~\ref{append:Expanded Data Pipeline} applies systematic filtering, score normalization, and benchmark alignment, model reports do not always disclose evaluation prompts, decoding settings, test splits, contamination controls, or metric variants with the same level of detail. As a result, BHI should not be viewed as replacing benchmark-specific replication studies. Instead, it provides a reproducible macro-level audit that can help identify where such more detailed studies are most needed.

The anti-saturation component also relies on several simplifying assumptions. In Section~\ref{sec:antisaturation}, we model short-term score evolution using a linear trend and a fixed projection window. This choice is intentionally conservative and is supported by the sensitivity analyses in Appendix~\ref{app:antisaturation_sensitivity}, but real benchmark lifecycles may be nonlinear: a benchmark can experience abrupt saturation after a new training recipe, tool-use capability, or data exposure event becomes widespread. Therefore, the dynamic component should be interpreted as a near-term warning signal rather than a long-horizon forecast.

The impact score in Section~\ref{sec:impact} measures capability-weighted adoption among models with valid public reports. This design avoids relying on raw popularity signals such as citations or repository statistics, but it also means that BHI captures reported evaluation practice rather than all forms of scientific influence. For example, a benchmark may shape research discussions, diagnostic analysis, or internal evaluation pipelines without appearing frequently in public model reports. Conversely, a widely reported benchmark may retain high adoption even after its technical diagnostic value has begun to decline. The three-axis decomposition is intended to expose this tension, but it cannot fully observe private or undocumented evaluation use.

Our external validation is also necessarily limited. The case study in Table~\ref{tab:validity} shows that BHI separates independently identified saturated benchmarks from active frontier benchmarks, providing evidence of directional consistency with current evaluation practice. However, this should be understood as face validity rather than criterion validity against an independently collected ground-truth health score. Building such a ground truth would require structured expert elicitation or longitudinal replication studies, which we leave for future work.

Finally, BHI is a lightweight post-hoc auditing procedure, but its scalability still depends on the availability and maintenance of structured benchmark metadata. The computational cost of the index itself is modest compared with model evaluation, since it operates on reported scores rather than rerunning benchmarks. The main cost lies in data curation, version alignment, and continuous updates as new models and benchmarks appear. Future work could integrate BHI with existing agent frameworks to support automated benchmark report parsing, metadata alignment, and periodic index updates as new models and benchmarks emerge.

\section{Expanded Related Work}
\label{appendix:expanded related work}

\subsection{Shifting Evaluation Paradigms}
The meteoric rise of LLMs has catalyzed a paradigm shift in evaluation, moving from single-task leaderboards toward comprehensive benchmark suites that encompass knowledge, reasoning, and safety \cite{liang2023holistic, srivastava2023beyond}. However, conventional evaluation frameworks predominantly adopt a model-centric perspective, treating benchmarks as static and immutable measurement tools. As model capabilities continue to leapfrog, public benchmarks are increasingly susceptible to score inflation and leaderboard distortion, which diminish their capacity to characterize genuine differences in model performance \cite{haimes2024benchmarkinflationrevealingllm}. Consequently, the research focus in the community is gradually pivoting from simple performance comparisons between models to in-depth investigations into the validity, robustness, and lifecycles of the benchmarks themselves \cite{heineman2025signal, zhou2025lost, qian2026_benchmark2}. Despite this shift, existing studies typically propose diagnostic metrics centered on local dimensions—such as uncertainty, measurement quality, or cross-benchmark consistency—and lack a unified, scalable framework for the longitudinal monitoring of benchmark efficacy as it co-evolves with model advancements.

\subsection{Reliability and Measurement Theory}
The reliability of evaluation outcomes faces severe challenges from prompt sensitivity, annotation noise, and stochasticity in sampling \cite{posix2024, madaan2024_variance}. To mitigate these issues, researchers have integrated statistical tools such as bootstrap estimation, confidence intervals, and significance testing to distinguish true performance gains from measurement noise, establishing more robust reporting norms \cite{dror2018hitchhikers}. From the perspectives of psychometrics and Item Response Theory (IRT), the core utility of a benchmark lies in its \textit{discriminative power}—the measurement resolution jointly determined by item difficulty and discrimination parameters \cite{lalor2016irt_scale}. In the context of LLM evaluation, however, high-performing models often trigger upper-tail compression and a lack of leaderboard separability, rendering traditional metrics ineffective at identifying fine-grained capability nuances.

\subsection{Benchmark Degradation and Dynamics}
The efficacy of evaluation benchmarks exhibits a discernible degradation trend over time, primarily driven by saturation effects and data contamination \cite{ott2022_mapping_benchmark_dynamics, DBLP:journals/corr/abs-2308-08493}. When models approach ceiling performance on existing datasets, or when evaluation data is implicitly leaked into pre-training corpora, the discriminative capacity of a benchmark shrinks rapidly. This blurs the boundary between generalization and memorization, prompting the development of contamination detection and verifiable evaluation corrections \cite{li-etal-2024-open-source, dong-etal-2024-generalization}. To address the loss of resolution caused by saturation, some work attempts to reactivate the discriminative power of existing benchmarks through weighted or augmented metrics, thus improving the separability among SOTA models \cite{etzine-etal-2025-revitalizing}. Although dynamic evaluation—which refreshes the task pools or introduces temporal windows—can alleviate contamination risks, it introduces a trade-off between maintaining longitudinal comparability and controlling evaluation costs \cite{kiela-etal-2021-dynabench, 10.1609/aaai.v38i17.29822}. Therefore, a significant research gap remains in the construction of a unified data-driven framework for the long-term tracking of benchmark health \cite{heineman2025signal, ott2022_mapping_benchmark_dynamics,ideabench10.1145/3711896.3737419}.

\section{Sensitivity Analysis of Capability Discrimination}
\label{app:threshold_sensitivity}

\subsection{EDR Threshold Sensitivity}
\label{app:edr_threshold_sensitivity}

The Effective Differentiation Ratio (EDR) in Section~3.1.1 depends on a noise-filtering threshold $\delta$, which determines whether a pairwise score gap is treated as an informative distinction between two models. In the main experiments, we set $\delta = 2.0\%$ of the observed score range for each benchmark. This appendix evaluates whether the resulting BHI ranking is sensitive to this choice.

We vary $\delta$ from $0.5\%$ to $5.0\%$ and recompute the complete BHI pipeline under each setting, including the normalized discrimination score, CRITIC-based final weighting, and global benchmark ranking. The resulting ranking is compared against the default setting $\delta=2.0\%$ using Spearman rank correlation and top-10 overlap.

\begin{table}[h]
\centering
\caption{Sensitivity of BHI rankings to the EDR threshold $\delta$. Spearman correlation is computed against the default setting $\delta=2.0\%$.}
\label{tab:threshold_sensitivity}
\begin{tabular}{cccccc}
\toprule
$\delta$ (\%) & Spearman & Top-10 Overlap & $w_{\mathrm{Disc}}$ & $w_{\mathrm{AS}}$ & $w_{\mathrm{Imp}}$ \\
\midrule
0.5 & 0.9582 & 10 & 0.3643 & 0.3386 & 0.2971 \\
1.0 & 0.9870 & 9  & 0.3255 & 0.3598 & 0.3147 \\
1.5 & 0.9963 & 9  & 0.3168 & 0.3644 & 0.3188 \\
2.0 & 1.0000 & 10 & 0.3263 & 0.3574 & 0.3163 \\
2.5 & 0.9955 & 10 & 0.3383 & 0.3465 & 0.3152 \\
3.0 & 0.9865 & 10 & 0.3603 & 0.3354 & 0.3043 \\
4.0 & 0.9764 & 9  & 0.3785 & 0.3265 & 0.2950 \\
5.0 & 0.9646 & 9  & 0.3914 & 0.3196 & 0.2890 \\
\bottomrule
\end{tabular}
\end{table}

As shown in Table~\ref{tab:threshold_sensitivity}, the BHI ranking remains highly stable across the tested threshold range. Even under the most conservative threshold setting of $\delta=5.0\%$, the Spearman correlation with the default ranking remains 0.9646, and the top-10 overlap remains 9 out of 10. The strongest stability is observed near the default region between $1.5\%$ and $2.5\%$, where the Spearman correlation exceeds 0.995.

\begin{figure}[h]
\centering
\includegraphics[width=0.75\linewidth]{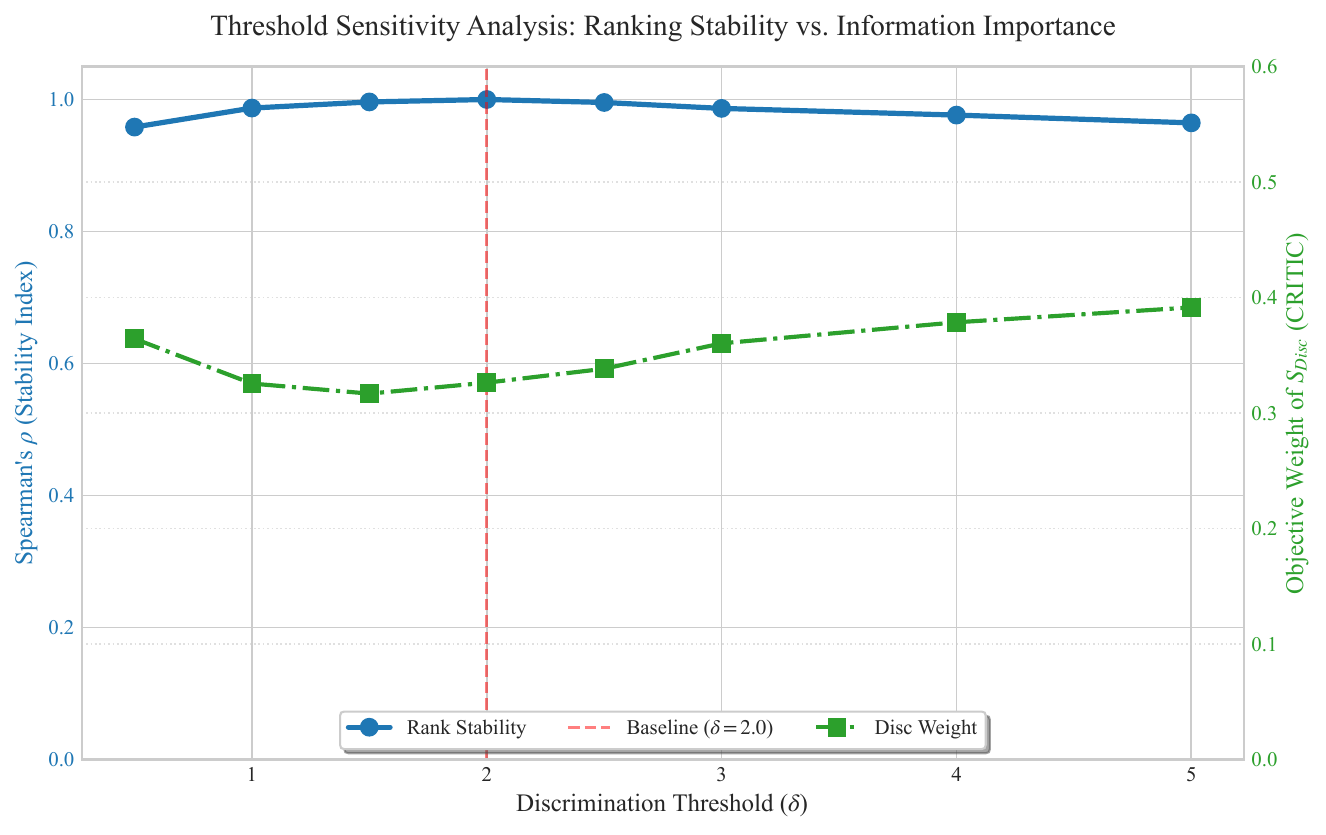}
\caption{Sensitivity of BHI ranking stability and the CRITIC-assigned discrimination weight under different EDR thresholds.}
\label{fig:threshold_sensitivity}
\end{figure}

Figure~\ref{fig:threshold_sensitivity} further shows that the CRITIC-assigned weight of $S_{\mathrm{Disc}}$ varies smoothly with $\delta$. Larger thresholds make EDR stricter and increase the relative information contribution of the discrimination axis, but this does not materially alter the final ranking. These results indicate that the proposed discrimination component captures a stable ranking signal rather than an artifact of a specific threshold choice.

\subsection{RCV Trimming Interval Sensitivity}
\label{app:rcv_sensitivity}

The RCV in Section~3.1.2 measures score spread using the P10--P90 interval, balancing robustness to extreme observations with retention of distributional information. To assess sensitivity to this choice, we compare six intervals from P0--P100 to P25--P75, including the default P10--P90. For each interval, we recompute the normalized discrimination score, CRITIC weights, and final BHI ranking, keeping the remaining settings unchanged.

\begin{table}[htbp]
\centering
\caption{Sensitivity of BHI rankings to the RCV percentile interval. All comparisons use P10--P90 as the reference. Rank shifts are absolute; weight drift is the maximum absolute change across the three CRITIC weights, measured in percentage points (pp).}
\label{tab:rcv_sensitivity}
\small
\setlength{\tabcolsep}{3pt}
\begin{tabular}{lccccccc}
\toprule
RCV interval & Spearman & Kendall & Top-10 & Top-20 & \multicolumn{2}{c}{Rank shift} & Max weight \\
 & $\rho$ & $\tau_b$ & overlap & overlap & Median & Max & drift (pp) \\
\midrule
P0--P100 & 0.9935 & 0.9463 & 9/10 & 17/20 & 2 & 29 & 1.806 \\
P5--P95 & 0.9978 & 0.9716 & 9/10 & 19/20 & 1 & 21 & 0.772 \\
\textbf{P10--P90} & \textbf{1.0000} & \textbf{1.0000} & \textbf{10/10} & \textbf{20/20} & \textbf{0} & \textbf{0} & \textbf{0} \\
P15--P85 & 0.9986 & 0.9740 & 10/10 & 18/20 & 1 & 9 & 0.293 \\
P20--P80 & 0.9962 & 0.9547 & 10/10 & 18/20 & 2 & 14 & 0.376 \\
P25--P75 & 0.9922 & 0.9374 & 9/10 & 18/20 & 3 & 25 & 1.321 \\
\bottomrule
\end{tabular}
\end{table}

As shown in Table~\ref{tab:rcv_sensitivity}, Spearman correlations remain at least 0.9922 across the tested intervals, top-10 overlap is at least 9 out of 10, and median absolute rank shifts do not exceed three positions. The maximum CRITIC-weight change is 1.806 percentage points. These results support the global and top-rank stability of BHI under the default P10--P90 interval and reasonable alternatives, rather than complete invariance of every individual rank.

\begin{figure*}[t]
    \centering
    \includegraphics[width=\textwidth]{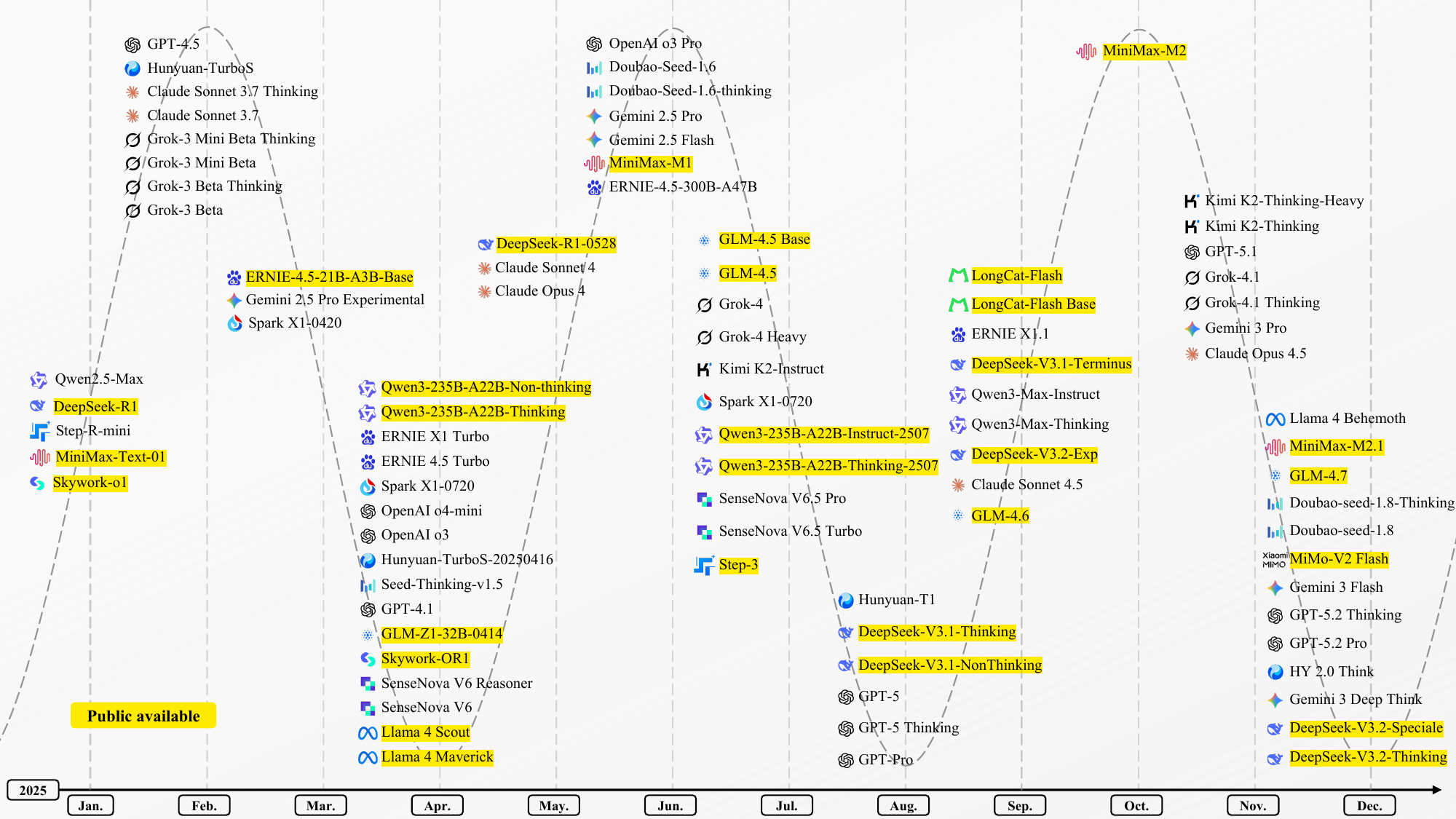} 
    \caption{Chronological distribution of the 91 mainstream large language models released throughout 2025.}
    \label{fig:model_timeline}
\end{figure*}

\section{SDM Weighting and Normalization}
\label{appendix:sdm_derivation}

This subsection provides the detailed mathematical derivation for SDM and the normalization process utilized in computing $S_{Disc}(b)$.

\subsection{Analysis of Metric Characteristics and Scale Disparities}
When constructing $S_{Disc}(b)$, the two constituent metrics, $EDR(b)$ and $RCV(b)$, exhibit significant differences in their mathematical properties and distributional behaviors:

\begin{enumerate}
    \item \textbf{Incongruent Scales and Units}: $RCV(b)$ is derived from the percentile spread of model score distributions and typically manifests as a continuous numerical value. In contrast, $EDR(b)$ represents the proportion of model pairs exceeding a specific noise-filtering threshold.
    \item \textbf{Variance Suppression Induced by Discretization}: The values of $EDR(b)$ are highly sensitive to the number of participating models $N$. For instance, if a benchmark has only $N=3$ models, the total number of pairwise combinations is $C_3^2 = 3$. Consequently, $EDR(b)$ can only take values from the discrete set $\{0, 1/3, 2/3, 1\}$. Such high discreteness in sparse datasets can lead to a "pseudo-standard deviation" that is statistically lower than the metric's true discriminative potential.
    \item \textbf{Sparse Coverage Bias}: Given that the study analyzes a meta-dataset of 158 benchmarks with varying model coverage, using raw standard deviations for weighting would allow metrics with naturally wider or more continuous distributions, such as $RCV(b)$, to dominate the final weight, thereby masking the fine-grained micro-resolution signals provided by $EDR(b)$.
\end{enumerate}

\subsection{Scale Alignment via Min-Max Normalization}
To mitigate the aforementioned biases, we first perform Min-Max normalization. For any indicator $k \in \{EDR, RCV\}$, the normalized value $Norm(k(b))$ is calculated as:

\begin{equation}
Norm(k(b)) = \frac{k(b) - \min(k)}{\max(k) - \min(k)}
\end{equation}

where $\min(k)$ and $\max(k)$ denote the minimum and maximum values of the metric across the entire set of 158 validated benchmarks. By mapping these indicators to the $[0, 1]$ interval, we effectively decouple the distributional shape from the magnitude scale. Within this normalized space, any observed fluctuation, represented by the standard deviation, solely reflects the contrast intensity of the metric in evaluating benchmark health, rather than being an artifact of the raw calculation formula or sampling density.

\subsection{Mathematical Derivation of SDM Weighting}
Within the normalized coordinate system, we employ SDM to determine the weights $w_k$, aiming to amplify signals with higher information load:

\begin{enumerate}
    \item \textbf{Calculation of Normalized Standard Deviation}:
    For each indicator $k \in \{EDR, RCV\}$, we compute its standard deviation $\sigma_k$ across all benchmarks:
    \begin{equation}
    \sigma_k = \sqrt{\frac{1}{M} \sum_{b=1}^{M} (Norm(k(b)) - \overline{Norm(k)})^2}
    \end{equation}
    where $M=158$ and $\overline{Norm(k)}$ denotes the mean value of the normalized metric in the sample set.

    \item \textbf{Weight Allocation}:
    The weight $w_k$ is determined by the proportion of the indicator's standard deviation relative to the total:
    \begin{equation}
    w_k = \frac{\sigma_k}{\sigma_{EDR} + \sigma_{RCV}}
    \end{equation}
    This mechanism ensures that if a metric remains more dispersed after normalization, it demonstrates higher sensitivity to the varying "health statuses" of different benchmarks and is thus assigned a higher weight.

\end{enumerate}

\section{Expanded Data Pipeline}
\label{append:Expanded Data Pipeline}
BHI employs a multi-stage pipeline to filter, standardize, and calibrate results from heterogeneous evaluation sets. As illustrated in Figure~\ref{fig:data_pipeline}, the workflow refines a broad corpus into a high-fidelity benchmark through three stages: collection, alignment, and validation.

\subsection{Data Collection}
We screened the official technical reports of 117 representative models released between 2025 and April 2026 to aggregate an initial corpus of 416 candidate benchmarks. Among them, 91 models were released in 2025, whose chronological distribution is shown in Figure~\ref{fig:model_timeline}. This selection covers both the 2025 model ecosystem and early-2026 frontier releases, providing sufficient coverage of generalized and specialized model capabilities.

The reports were identified through targeted searches and manual screening. Scores from clearly structured tables were extracted using rule-based methods, with manual review for heterogeneous layouts and ambiguous records. Before inclusion in the score matrix, extracted entries were checked against the original reports. These checks covered benchmark names and versions, metric direction, evaluation settings where reported, duplicate entries, and the distinction between missing values and reported zero scores, helping reduce inconsistencies before the subsequent alignment stage.

\subsection{Data Alignment}
To ensure comparability across disparate metrics and varying evaluation conditions, we implement a multi-stage alignment protocol. This process standardizes raw scores and addresses the structured requirements of different benchmarks.

\subsubsection{Filtering, Normalization and calibration} 
We apply a \textit{Minimum Participation Threshold} to exclude benchmarks evaluated by fewer than three models, ensuring the statistical validity of our dispersion-based metrics. Subsequently, all scores are normalized to a $[0, 100]$ scale. For cost-oriented metrics where lower values indicate better performance, we apply a subtraction-based transformation to align them with a standard reward-centric orientation.

For open-ended metrics lacking a theoretical maximum—such as task-specific throughput or reward-modeling scores—we apply a \emph{Headroom Buffer Factor} of $1.25$. The normalized score $S_{\mathrm{norm}}$ is calculated as:
\begin{equation}
S_{\mathrm{norm}} = \frac{S_{\mathrm{raw}}}{1.25\, S_{\max,\mathrm{obs}}} \times 100
\end{equation}
where $S_{\max,\mathrm{obs}}$ represents the maximum observed performance within the dataset. As illustrated in the \emph{Calibration} step of Figure~\ref{fig:data_pipeline}(a), this maps the current SOTA performance to 80 points, reserving a 20\% headroom buffer. The factor 1.25 is an operational choice rather than a theoretically optimal value. To assess its sensitivity, we rerun the complete BHI pipeline with alternative factors of 1.10, 1.20, 1.33, and 1.50. Relative to the default setting, all alternatives yield Spearman correlations of at least 0.9978, retain at least 19 of the top-20 benchmarks, and produce median absolute rank changes of at most one position. The six eligible open-ended benchmarks also preserve their internal ordering. These results indicate that the overall ranking is insensitive to the exact buffer factor within the tested range.

\subsubsection{Specialized Handling} 
For temporal datasets such as LiveCodeBench, we aggregated fragmented, time-bound snapshots into major version clusters under the assumption of difficulty invariance. This consolidation maintains a dense evaluation matrix while preserving the temporal integrity of the data. For tool-augmented evaluations, we implemented a bifurcated selection policy: we utilized base-ability scores for Capability Discrimination to isolate the benchmark's intrinsic resolution in separating model architectures, while incorporating all performance gains—including agentic and retrieval-based optimizations—for Anti-Saturation to evaluate the practical exhaustion of the task environment.

\subsection{Validated Benchmark Set}
After rigorous alignment and filtering, we retained 158 validated benchmarks. The domain distribution (Figure~\ref{fig:data_pipeline}(b)) highlights BHI's diverse evaluation dimensions, covering core domains like Reasoning (20.0\%) and Multimodal (14.8\%) alongside Mathematics and Code generation. Moreover, the vendor distribution (Figure~\ref{fig:data_pipeline}(c)) reflects ecosystem diversity, spanning closed-source (OpenAI, Anthropic), open-source (DeepSeek, Alibaba), and domain-specific models (Meituan, Xiaomi). This curated suite establishes a high-fidelity foundation for statistical analysis and BHI health metric computation.

\section{CRITIC Objective Weighting Method}
\label{appendix:critic_derivation}

This section provides the detailed mathematical derivation for the CRITIC method used to synthesize the final BHI from the three primary dimensions: $S_{Disc}$, $S_{AS}$, and $S_{Imp}$.

\subsection{Statistical Logic of the CRITIC Method}
The CRITIC method is an objective weighting approach that evaluates the importance of each dimension by considering both its contrast intensity and the conflicting character between indicators. Unlike simpler methods like the entropy weight method, CRITIC accounts for the correlations between variables, ensuring that the weights reflect the total independent information volume carried by each dimension within the global distribution of the benchmarks.

\subsection{Mathematical Derivation Steps}
Let $M$ denote the total number of benchmarks and $n$ denote the number of evaluation dimensions ($n=3$).

\begin{enumerate}
    \item \textbf{Dimensional Normalization}: 
    Since the calculation logics for $S_{Disc}$, $S_{AS}$, and $S_{Imp}$ differ, all input dimensions are first mapped to the $[0, 1]$ interval via Min-Max normalization to eliminate the influence of scales and units.

    \item \textbf{Calculation of Contrast Intensity}: 
    The contrast intensity of the $j$-th dimension is represented by its standard deviation $\sigma_j$. A higher standard deviation indicates a greater variation in values across different benchmarks, suggesting that the dimension provides more discriminative information.
    \begin{equation}
    \sigma_j = \sqrt{\frac{1}{M-1} \sum_{b=1}^M (x_{bj} - \bar{x}_j)^2}
    \end{equation}
    where $x_{bj}$ is the normalized score of benchmark $b$ on dimension $j$, and $\bar{x}_j$ is the mean value.

    \item \textbf{Calculation of Conflicting Character}: 
    The conflicting character represents the degree of correlation between indicators. Let $r_{jk}$ denote the Pearson correlation coefficient between the $j$-th and $k$-th dimensions. If a dimension is highly correlated with others, it contains more redundant information, and its independent weight should be adjusted accordingly. The conflict index $R_j$ is defined as:
    \begin{equation}
    R_j = \sum_{k=1}^n (1 - r_{jk})
    \end{equation}

    \item \textbf{Calculation of Information Volume}: 
    The comprehensive information volume $C_j$ for the $j$-th dimension is the product of its contrast intensity and its conflicting character. A larger $C_j$ value indicates that the dimension carries more total information within the evaluation system.
    \begin{equation}
    C_j = \sigma_j \cdot R_j = \sigma_j \sum_{k=1}^n (1 - r_{jk})
    \end{equation}

    \item \textbf{Determination of Final Weights}: 
    The final objective weights $w_j$ are obtained by normalizing the $C_j$ values:
    \begin{equation}
    w_j = \frac{C_j}{\sum_{k=1}^n C_k}
    \end{equation}
\end{enumerate}

\section{Robustness under Score Noise and Model Dropout}
\label{app:robustness_noise_dropout}

This appendix evaluates the robustness of BHI under two common sources of instability in large-scale benchmark meta-analysis: score perturbation and incomplete model coverage. The first experiment injects controlled Gaussian noise into benchmark scores, simulating reporting variation, measurement uncertainty, and minor evaluation inconsistencies. The second experiment randomly removes evaluated models, simulating incomplete reporting and partial benchmark coverage.

For both experiments, we rerun the full BHI computation pipeline after perturbation, including score normalization, LOBO-based model capability calibration, component construction, and final aggregation. We compare the perturbed rankings with the original baseline ranking using Spearman rank correlation, Kendall rank correlation, top-$k$ overlap, and mean rank shift. To evaluate whether the CRITIC weighting scheme improves stability, we also report results under an equal-weight aggregation baseline.

\subsection{Robustness to Gaussian Score Noise}
\label{app:noise_robustness}

We inject Gaussian noise into the normalized benchmark scores under different noise levels $\sigma \in \{0.01, 0.02, 0.05, 0.10, 0.15, 0.20\}$. Scores are clipped to the valid interval after perturbation. For each noise level, the complete BHI pipeline is recomputed over repeated trials, and the resulting rankings are compared with the clean baseline.

\begin{table}[h]
\centering
\caption{Noise robustness under CRITIC weighting. Score stability is measured as $1-\mathrm{MAE}$ between perturbed and baseline BHI scores.}
\label{tab:noise_critic}
\begin{tabular}{cccccc}
\toprule
$\sigma$ & Spearman & Kendall & Score Stability & Top-10 Overlap & Mean Rank Shift \\
\midrule
0.01 & 0.9911 & 0.9353 & 0.9898 & 9.67 & 3.70 \\
0.02 & 0.9827 & 0.9027 & 0.9852 & 9.45 & 5.53 \\
0.05 & 0.9544 & 0.8329 & 0.9755 & 8.90 & 9.44 \\
0.10 & 0.9184 & 0.7674 & 0.9650 & 8.67 & 13.00 \\
0.15 & 0.8854 & 0.7189 & 0.9543 & 8.44 & 15.69 \\
0.20 & 0.8575 & 0.6804 & 0.9458 & 8.36 & 17.75 \\
\bottomrule
\end{tabular}
\end{table}

\begin{table}[h]
\centering
\caption{Noise robustness under equal weighting.}
\label{tab:noise_equal}
\begin{tabular}{cccccc}
\toprule
$\sigma$ & Spearman & Kendall & Score Stability & Top-10 Overlap & Mean Rank Shift \\
\midrule
0.01 & 0.9909 & 0.9306 & 0.9892 & 9.19 & 3.99 \\
0.02 & 0.9819 & 0.8967 & 0.9845 & 9.14 & 5.91 \\
0.05 & 0.9523 & 0.8258 & 0.9744 & 8.78 & 9.89 \\
0.10 & 0.9106 & 0.7527 & 0.9619 & 8.36 & 13.81 \\
0.15 & 0.8748 & 0.7013 & 0.9504 & 7.93 & 16.67 \\
0.20 & 0.8425 & 0.6595 & 0.9405 & 7.72 & 18.90 \\
\bottomrule
\end{tabular}
\end{table}

Tables~\ref{tab:noise_critic} and~\ref{tab:noise_equal} show that BHI rankings remain stable under moderate score perturbations. Under CRITIC weighting, the Spearman correlation remains above 0.95 when $\sigma \leq 0.05$ and above 0.90 when $\sigma \leq 0.10$. Even under the largest tested perturbation, $\sigma=0.20$, the Spearman correlation remains 0.8575 and the BHI score stability remains 0.9458. This suggests that the final ranking is not dominated by small score-level fluctuations.

Compared with equal weighting, CRITIC weighting yields consistently higher Spearman correlation, Kendall correlation, score stability, and top-10 overlap across most noise levels. The gap becomes more visible as the noise level increases. This supports the use of CRITIC as a data-driven aggregation mechanism, because it assigns lower influence to less informative or more redundant components rather than treating all dimensions as equally reliable under perturbation.

\begin{figure}[h]
\centering
\includegraphics[width=0.72\linewidth]{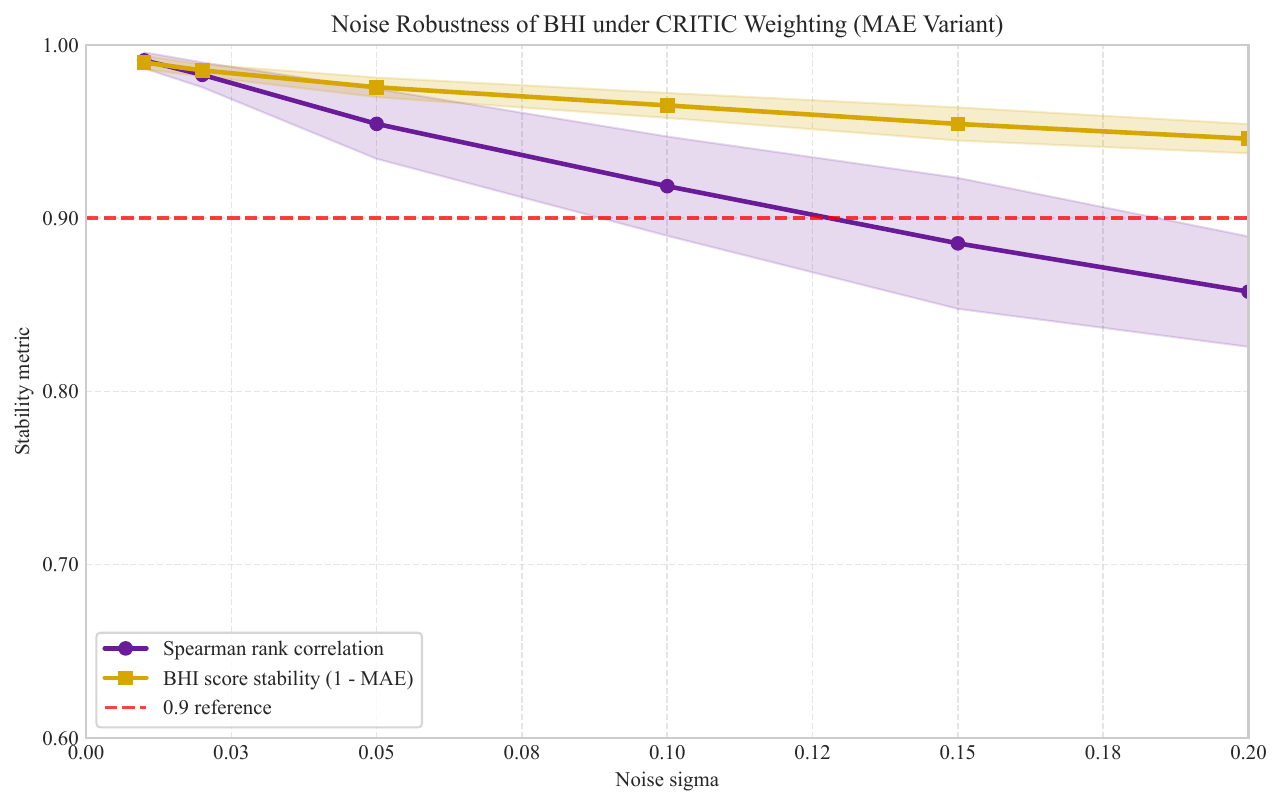}
\caption{Noise robustness of BHI under CRITIC weighting.}
\label{fig:noise_critic}
\end{figure}

\begin{figure}[h]
\centering
\includegraphics[width=0.72\linewidth]{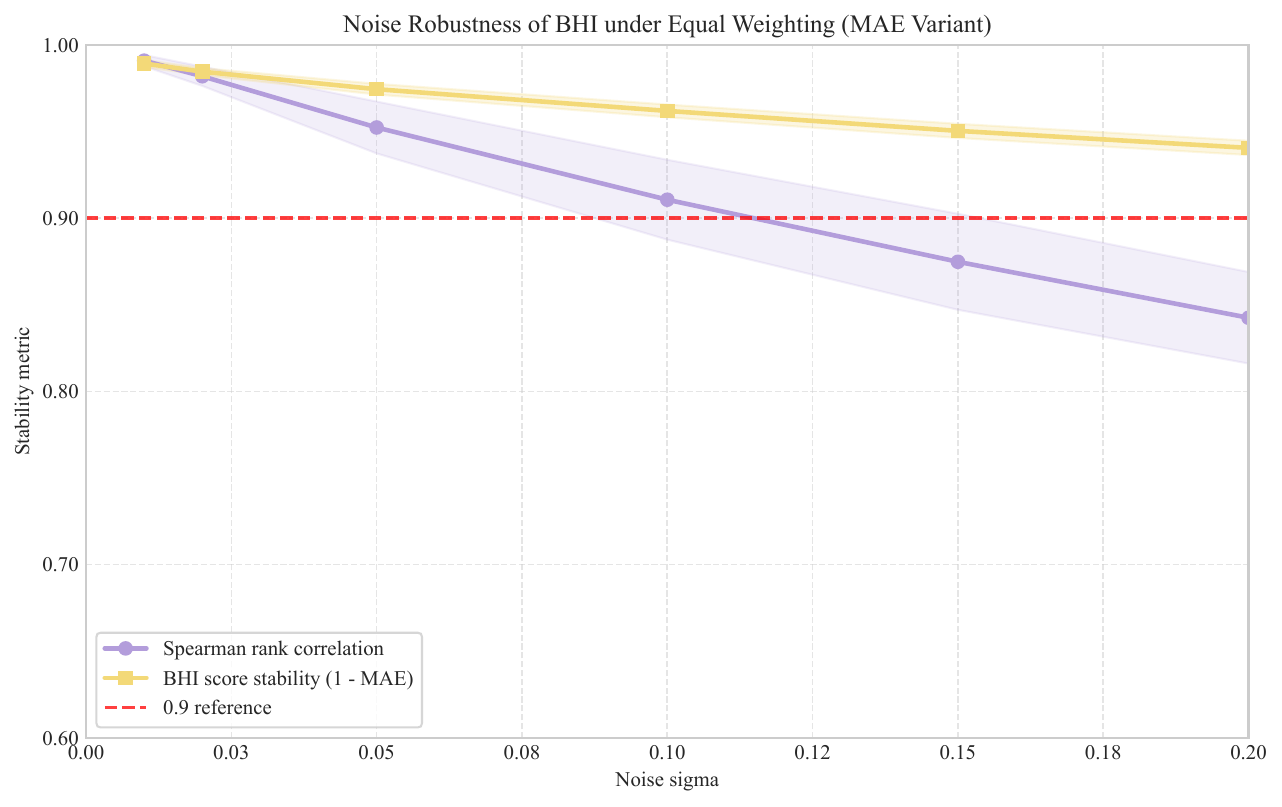}
\caption{Noise robustness of BHI under equal weighting.}
\label{fig:noise_equal}
\end{figure}

Figures~\ref{fig:noise_critic} and~\ref{fig:noise_equal} visualize the decay of ranking stability as the perturbation magnitude increases. The degradation is gradual rather than abrupt, indicating that the BHI ranking is robust to realistic levels of measurement noise.

\subsection{Robustness to Model Dropout}
\label{app:model_dropout}

We further evaluate robustness under incomplete model coverage by randomly removing a fraction of models from the meta-dataset. The dropout ratio ranges from 5\% to 60\%. After each removal, all benchmark-level metrics are recomputed from the remaining models. Benchmarks without sufficient remaining observations are excluded from that trial, and the final ranking is compared against the original baseline on the common valid set.

\begin{table}[h]
\centering
\caption{Model dropout robustness under CRITIC weighting.}
\label{tab:dropout_critic}
\begin{tabular}{cccccc}
\toprule
Dropout Ratio & Spearman & Kendall & Top-10 Overlap & Mean Rank Shift & Models Used \\
\midrule
0.05 & 0.9961 & 0.9667 & 9.31 & 2.99 & 112 \\
0.10 & 0.9911 & 0.9389 & 9.14 & 5.12 & 106 \\
0.15 & 0.9851 & 0.9134 & 9.04 & 7.56 & 100 \\
0.20 & 0.9814 & 0.8986 & 8.91 & 9.31 & 94 \\
0.25 & 0.9769 & 0.8835 & 8.75 & 11.34 & 88 \\
0.30 & 0.9716 & 0.8684 & 8.48 & 13.53 & 82 \\
0.35 & 0.9646 & 0.8509 & 8.47 & 14.61 & 77 \\
0.40 & 0.9591 & 0.8388 & 8.29 & 17.09 & 71 \\
0.45 & 0.9549 & 0.8285 & 8.25 & 18.99 & 65 \\
0.50 & 0.9474 & 0.8140 & 7.86 & 22.11 & 59 \\
0.55 & 0.9343 & 0.7923 & 7.74 & 23.76 & 53 \\
0.60 & 0.9264 & 0.7791 & 7.55 & 26.51 & 47 \\
\bottomrule
\end{tabular}
\end{table}

\begin{table}[h]
\centering
\caption{Model dropout robustness under equal weighting.}
\label{tab:dropout_equal}
\begin{tabular}{cccccc}
\toprule
Dropout Ratio & Spearman & Kendall & Top-10 Overlap & Mean Rank Shift & Models Used \\
\midrule
0.05 & 0.9956 & 0.9586 & 8.83 & 3.30 & 112 \\
0.10 & 0.9909 & 0.9353 & 8.90 & 5.20 & 106 \\
0.15 & 0.9854 & 0.9131 & 8.81 & 7.57 & 100 \\
0.20 & 0.9815 & 0.8984 & 8.75 & 9.32 & 94 \\
0.25 & 0.9768 & 0.8833 & 8.69 & 11.37 & 88 \\
0.30 & 0.9717 & 0.8692 & 8.55 & 13.55 & 82 \\
0.35 & 0.9652 & 0.8528 & 8.63 & 14.58 & 77 \\
0.40 & 0.9600 & 0.8410 & 8.49 & 17.08 & 71 \\
0.45 & 0.9554 & 0.8303 & 8.46 & 18.99 & 65 \\
0.50 & 0.9479 & 0.8152 & 8.03 & 22.13 & 59 \\
0.55 & 0.9367 & 0.7967 & 8.16 & 23.70 & 53 \\
0.60 & 0.9301 & 0.7856 & 7.86 & 26.45 & 47 \\
\bottomrule
\end{tabular}
\end{table}

Tables~\ref{tab:dropout_critic} and~\ref{tab:dropout_equal} show that the BHI ranking is robust to substantial model removal. Under CRITIC weighting, the Spearman correlation remains above 0.99 at 10\% dropout, above 0.97 up to 30\% dropout, and above 0.92 even when 60\% of models are removed. The top-10 overlap also degrades gradually, indicating that the highest-ranked benchmarks are not highly dependent on a small subset of model reports.

\begin{figure}[h]
\centering
\includegraphics[width=0.72\linewidth]{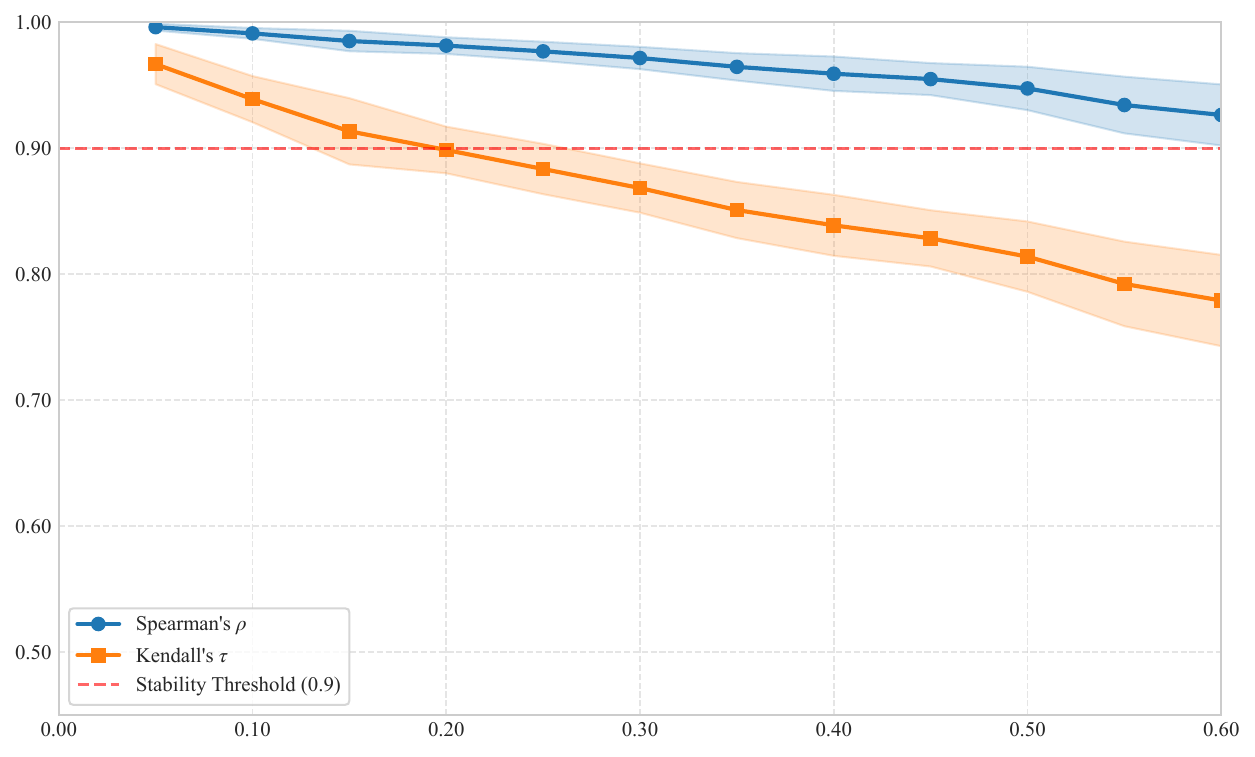}
\caption{Ranking stability under model dropout using CRITIC weighting.}
\label{fig:dropout_critic}
\end{figure}

\begin{figure}[h]
\centering
\includegraphics[width=0.72\linewidth]{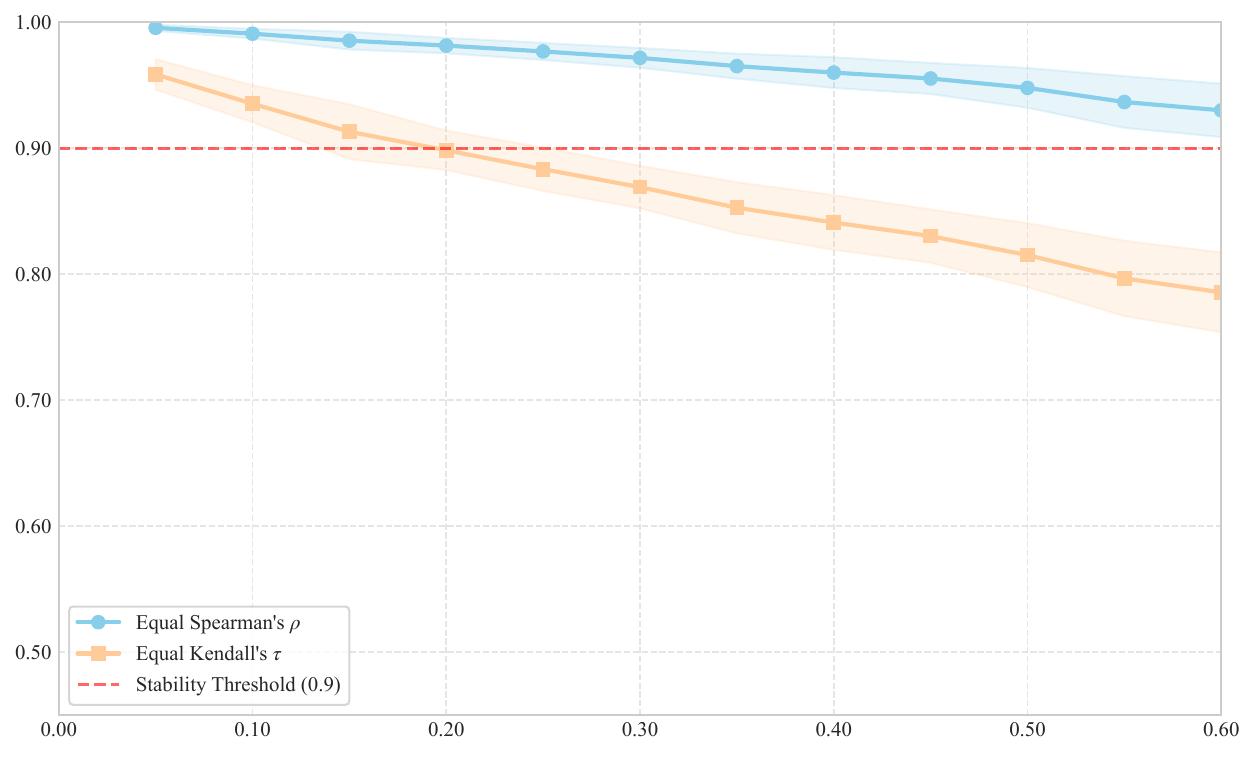}
\caption{Ranking stability under model dropout using equal weighting.}
\label{fig:dropout_equal}
\end{figure}

The comparison between CRITIC and equal weighting indicates that both aggregation schemes are stable under model dropout, while CRITIC generally provides slightly stronger top-rank preservation in the low-to-moderate dropout regime. This is consistent with the role of CRITIC weighting as an objective mechanism that adjusts component contributions according to their information content and independence.

\subsection{Robustness to Targeted Score Errors}
\label{app:targeted_score_errors}

To complement the Gaussian-noise analysis, we examine targeted downward score errors on the ten benchmarks with the lowest static Anti-Saturation scores. For each benchmark, we identify the lowest-capability 10\% of evaluated models using target-excluded LOBO estimates and reduce their normalized scores by 10, 15, 20, 25, or 30 points, clipping at zero. We repeat the same perturbations for the highest-capability 10\% as a comparison and recompute the complete BHI pipeline, yielding 100 runs.

At the 30-point reduction, low-capability perturbations yield median Spearman and Kendall correlations of 0.9982 and 0.9931, respectively. The median direct change in $S_{\mathrm{AS}}$ is 0.0333 for low-capability models, compared with 0.0907 for high-capability models. This comparison suggests that LOBO-based capability weighting moderates the influence of low-capability score errors on Anti-Saturation. Together with the high ranking correlations, these results indicate that BHI can accommodate the tested score perturbations while largely preserving the overall benchmark ordering.

\section{Sensitivity Analysis of Anti-Saturation Parameters}
\label{app:antisaturation_sensitivity}

The Anti-Saturation score $S_{\mathrm{AS}}$ combines a static resistance component $S_{\mathrm{Sta}}$ and a dynamic saturation projection component $S_{\mathrm{Dyn}}$. In the main experiments, we use a 30-day projection window and combine the two components as
\[
S_{\mathrm{AS}} = 0.8 \cdot S_{\mathrm{Sta}} + 0.2 \cdot S_{\mathrm{Dyn}}.
\]
This appendix evaluates whether the final BHI ranking is sensitive to these two design choices.

\subsection{Sensitivity to the Dynamic Projection Window}
\label{app:dynamic_window}

We vary the temporal projection window $T$ from 20 to 40 days and recompute the dynamic saturation score as
\[
S_{\mathrm{Dyn}}(b) = \max(0, 1 - (\overline{\mathrm{Score}}_b + T \cdot k)).
\]
The default setting corresponds to $T=30$ days. For each alternative window size, we recompute the full BHI ranking and compare it with the default ranking.

\begin{table}[h]
\centering
\caption{Sensitivity of BHI rankings to the temporal projection window in $S_{\mathrm{Dyn}}$.}
\label{tab:dynamic_window}
\begin{tabular}{cccccc}
\toprule
Window $T$ & Spearman & Kendall & Top-10 Overlap & Top-20 Overlap & Avg. Rank Shift \\
\midrule
20 & 0.9999 & 0.9966 & 10 & 20 & 0.2405 \\
25 & 1.0000 & 0.9982 & 10 & 20 & 0.1392 \\
30 & 1.0000 & 1.0000 & 10 & 20 & 0.0000 \\
35 & 0.9999 & 0.9973 & 10 & 20 & 0.2152 \\
40 & 0.9999 & 0.9955 & 10 & 20 & 0.3291 \\
\bottomrule
\end{tabular}
\end{table}

Table~\ref{tab:dynamic_window} shows that the BHI ranking is nearly invariant to the temporal projection window. Across all tested values, the Spearman correlation remains above 0.9998, the top-10 and top-20 sets are fully preserved, and the average rank shift remains below 0.33. This indicates that the dynamic component captures only a smooth short-term saturation adjustment and does not introduce instability into the final ranking.

\begin{figure}[h]
\centering
\includegraphics[width=0.72\linewidth]{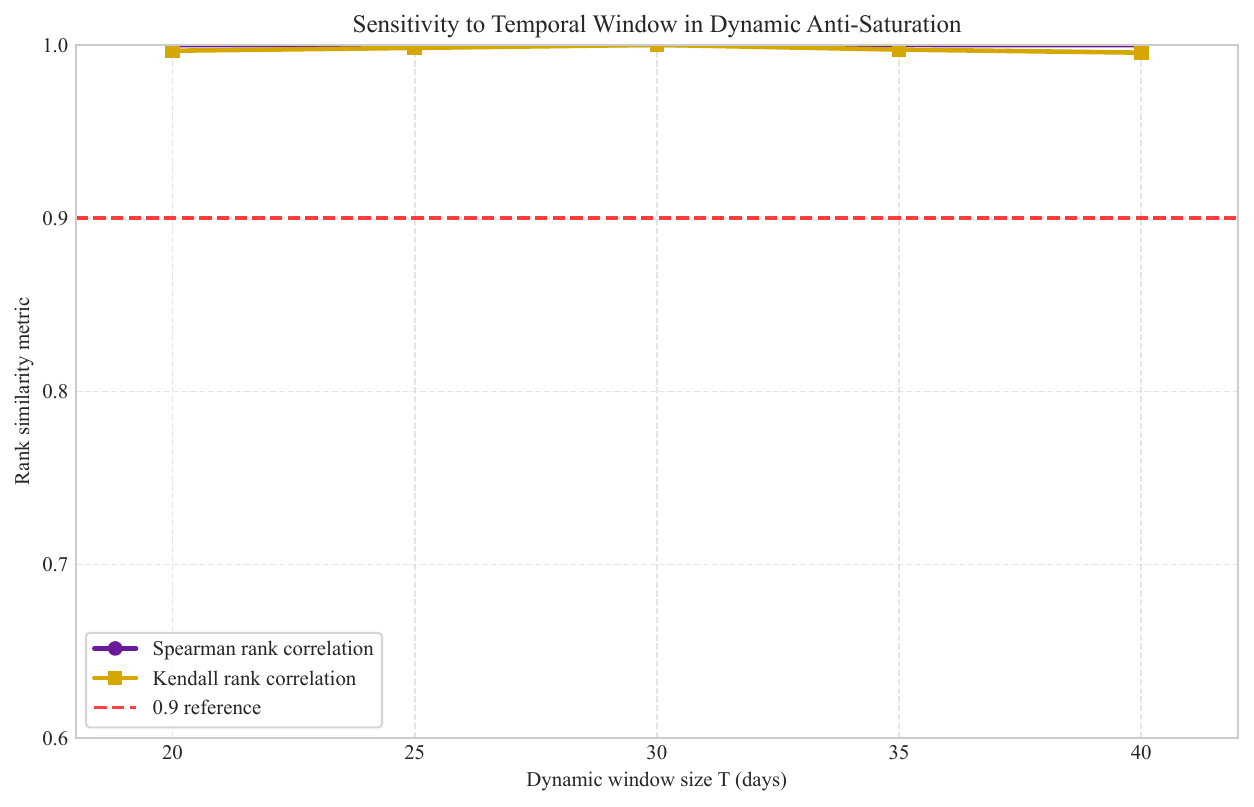}
\caption{Sensitivity of BHI rankings to the temporal window used in the dynamic anti-saturation component.}
\label{fig:dynamic_window}
\end{figure}

\subsection{Sensitivity to Static--Dynamic Weighting}
\label{app:static_dynamic_weight}

We next vary the relative contribution of $S_{\mathrm{Sta}}$ and $S_{\mathrm{Dyn}}$ in the final anti-saturation score. The default setting assigns 80\% weight to the static component and 20\% weight to the dynamic component. We compare this setting with alternative static weights from 0.6 to 0.9.

\begin{table}[h]
\centering
\caption{Sensitivity of BHI rankings to the static--dynamic weighting in $S_{\mathrm{AS}}$.}
\label{tab:static_dynamic_weight}
\begin{tabular}{ccccccc}
\toprule
$w_{\mathrm{Sta}}$ & $w_{\mathrm{Dyn}}$ & Spearman & Kendall & Top-10 Overlap & Top-20 Overlap & Avg. Rank Shift \\
\midrule
0.6 & 0.4 & 0.9995 & 0.9889 & 10 & 20 & 0.7722 \\
0.7 & 0.3 & 0.9998 & 0.9931 & 10 & 20 & 0.5063 \\
0.8 & 0.2 & 1.0000 & 1.0000 & 10 & 20 & 0.0000 \\
0.9 & 0.1 & 0.9998 & 0.9939 & 9  & 20 & 0.4177 \\
\bottomrule
\end{tabular}
\end{table}

As shown in Table~\ref{tab:static_dynamic_weight}, the final ranking is also robust to the static--dynamic weighting ratio. All tested configurations yield Spearman correlations above 0.999, and the top-20 benchmarks are fully preserved. The largest observed average rank shift is below one position, indicating that the anti-saturation axis is primarily driven by persistent benchmark headroom rather than by short-term trend extrapolation.

\begin{figure}[h]
\centering
\includegraphics[width=0.72\linewidth]{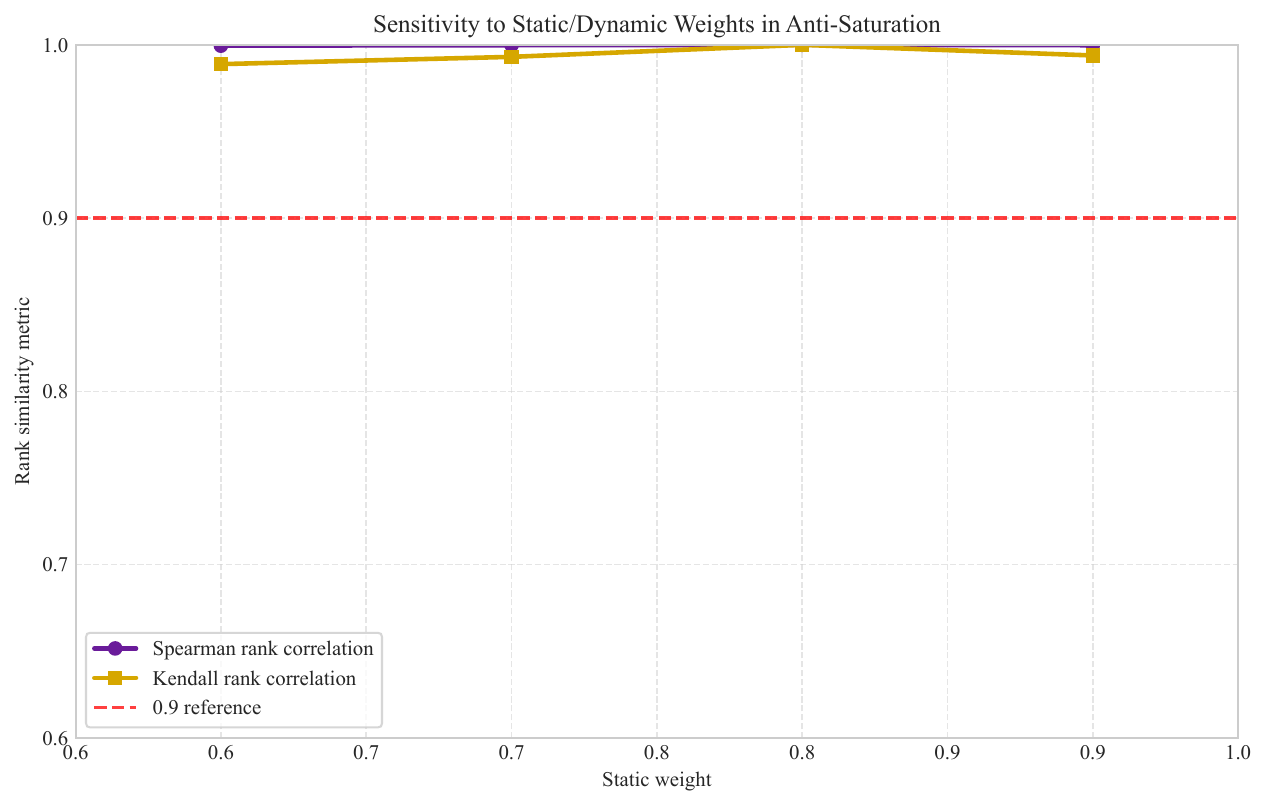}
\caption{Sensitivity of BHI rankings to the relative weighting between static resistance and dynamic saturation projection.}
\label{fig:static_dynamic_weight}
\end{figure}

Overall, these results support the use of a predominantly static formulation for $S_{\mathrm{AS}}$, while retaining a smaller dynamic term to reflect short-term score inflation. The selected 0.8/0.2 configuration provides a conservative balance between present benchmark difficulty and near-future saturation risk.

\section{Coverage and Cold-Start Bias Analysis}
\label{app:coverage_bias}

A potential concern for any benchmark-level meta-index is that highly evaluated benchmarks may receive higher scores simply because they have broader reporting coverage. This concern is particularly relevant for BHI because the impact component $S_{\mathrm{Imp}}$ is designed to measure capability-weighted adoption. We therefore analyze whether the final BHI ranking is dominated by benchmark coverage.

For each benchmark, we define coverage as the number of models with a valid reported score on that benchmark. We then compute Pearson and Spearman correlations between coverage and each BHI component, as well as the final BHI score.

\begin{table}[h]
\centering
\caption{Correlation between benchmark coverage and BHI components.}
\label{tab:coverage_correlation}
\begin{tabular}{ccc}
\toprule
Metric & Pearson Correlation & Spearman Correlation \\
\midrule
$S_{\mathrm{Disc}}$ & 0.2807 & 0.2424 \\
$S_{\mathrm{AS}}$   & -0.0275 & -0.0333 \\
$S_{\mathrm{Imp}}$  & 0.9130 & 0.8862 \\
BHI                 & 0.4737 & 0.4000 \\
\bottomrule
\end{tabular}
\end{table}

Table~\ref{tab:coverage_correlation} shows that coverage is strongly correlated with $S_{\mathrm{Imp}}$, which is expected by construction because $S_{\mathrm{Imp}}$ measures capability-weighted adoption. However, coverage has only weak correlation with $S_{\mathrm{Disc}}$ and almost no correlation with $S_{\mathrm{AS}}$. This indicates that broader reporting does not automatically imply stronger discriminative power or greater resistance to saturation.

The final BHI score exhibits only moderate correlation with coverage, with Pearson correlation 0.4737 and Spearman correlation 0.4000. This suggests that benchmark coverage contributes to the final score through the impact dimension, but does not dominate the overall ranking. In particular, a benchmark must still provide discriminative resolution and saturation resistance to receive a high BHI score.

\begin{figure}[h]
\centering
\includegraphics[width=0.72\linewidth]{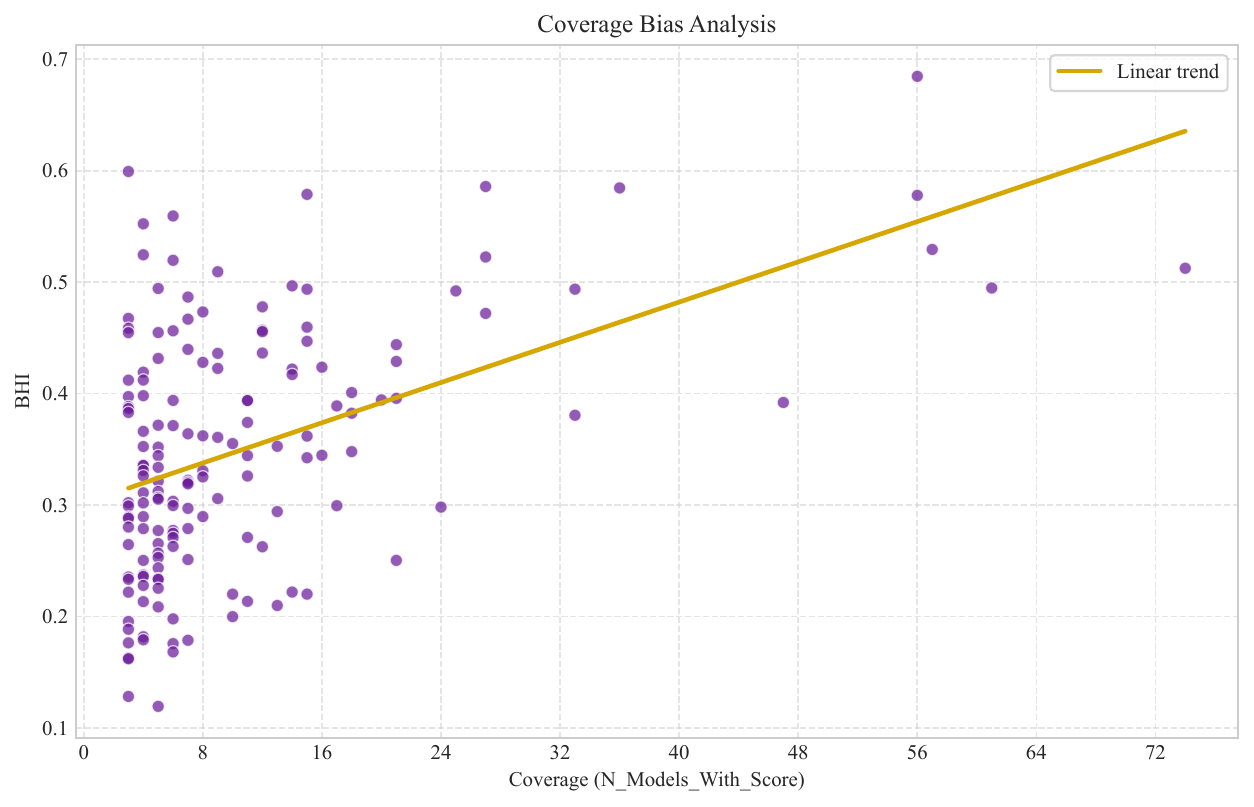}
\caption{Relationship between benchmark coverage and final BHI score. Coverage is measured as the number of models with valid scores on each benchmark.}
\label{fig:coverage_vs_bhi}
\end{figure}

Figure~\ref{fig:coverage_vs_bhi} provides a complementary visualization. Although the fitted trend line is positive, the scatter distribution is broad: benchmarks with similar coverage can receive substantially different BHI scores, and several low-coverage benchmarks obtain competitive BHI values when they exhibit strong discrimination or anti-saturation properties. Conversely, some widely reported benchmarks receive relatively low BHI scores due to saturation or reduced discriminative capacity.

These results confirm that BHI is not a proxy for benchmark popularity or reporting frequency. Instead, the final score reflects a balance among technical resolution, remaining headroom, and capability-weighted adoption.

Moreover, we examine the relationships between benchmark age, model adoption, and BHI scores to assess whether cold start systematically penalizes newly released benchmarks. Using April 30, 2026 as the reference date, we compute benchmark age from verified first public release dates and analyze its associations with model coverage, $S_{\mathrm{Imp}}$, and BHI, using 10,000 bootstrap resamples. The respective Spearman correlations are 0.020, $-0.271$, and $-0.440$. Among benchmarks released within 12 months, 36.6\% enter the top BHI quartile, compared with 21.6\% of older benchmarks.

Within the six-month subset, age correlates positively with coverage ($\rho = 0.617$) and BHI ($\rho = 0.704$), suggesting an initial period of adoption accumulation. These associations become less consistent across the 9-, 12-, and 18-month windows. Taken together, the results suggest that early adoption accumulation does not translate into a persistent ranking disadvantage for newer benchmarks in the observed sample. Their representation in the top quartile further indicates that recently released benchmarks can attain competitive BHI scores alongside established evaluation sets.

\section{Expanded Case Study}
\label{append:doman-specific-analyisis}

To illustrate how BHI supports benchmark-level diagnosis rather than merely producing a scalar ranking, we analyze three representative cases with distinct metric profiles: Humanity's Last Exam as the strongest overall frontier benchmark in the Knowledge QA category, SimpleQA as a high-resolution short-form factuality benchmark, and ZeroBench as a reasoning benchmark with limited current adoption but substantial remaining headroom. These cases show that the three BHI dimensions capture different types of benchmark value and potential limitations in the current evaluation ecosystem.

\paragraph{Humanity's Last Exam.}
Humanity's Last Exam (HLE) \cite{phan2025humanitysexam} ranks first in our benchmark audit with a BHI score of $0.6846$. Its advantage does not come from an extreme value on a single axis, but from a balanced and consistently strong profile across discrimination, saturation resistance, and impact. Specifically, HLE achieves $S_{\mathrm{Disc}} = 0.6608$, $S_{\mathrm{Sat}} = 0.6299$, and $S_{\mathrm{Imp}} = 0.7710$, indicating that it can separate current models effectively, preserve sufficient remaining challenge space, and maintain broad ecosystem recognition. This makes HLE a suitable anchor for tracking frontier progress in knowledge-intensive and reasoning-oriented evaluation: it combines technical difficulty with stable and influential reporting value in the broader evaluation ecosystem.

\paragraph{SimpleQA.}
SimpleQA \cite{wei2024measuringshortformfactualitylarge} obtains a BHI score of $0.5857$, reflecting the distinctive role of short-form factuality evaluation in the current benchmark landscape. Its strongest advantage lies in Capability Discrimination, where it reaches $S_{\mathrm{Disc}} = 0.8130$, the highest among the three cases. This suggests that SimpleQA's concise and unambiguous question design reduces interference from complex linguistic formulation and multi-step reasoning, thereby isolating differences in factual answering ability more effectively. At the same time, its saturation score remains moderate ($S_{\mathrm{Sat}} = 0.5043$), indicating that it still preserves useful challenge space, though its long-term frontier-tracking capacity is more limited than that of HLE or ZeroBench. Its impact score ($S_{\mathrm{Imp}} = 0.4435$) further suggests that SimpleQA has gained meaningful adoption, but has not yet reached the ecosystem coverage of broader frontier benchmarks such as HLE. Therefore, SimpleQA is best viewed as a high-resolution diagnostic tool for factuality evaluation rather than as a standalone benchmark for long-term frontier capability tracking.

\paragraph{ZeroBench.}
ZeroBench \cite{roberts2025zerobenchimpossiblevisualbenchmark} exhibits a different metric structure. Although its impact score is limited ($S_{\mathrm{Imp}} = 0.2498$), BHI assigns it a relatively high overall score of $0.5593$ due to its strong structural health. Its saturation score reaches $S_{\mathrm{Sat}} = 0.8687$, the highest among the three cases, indicating that current models remain far from exhausting its challenge space. Meanwhile, its discrimination score ($S_{\mathrm{Disc}} = 0.5205$) suggests that the benchmark has not collapsed into uniform failure, but can still produce meaningful performance variation across models. This profile identifies ZeroBench as a high-headroom reasoning benchmark whose current adoption underestimates its technical value. For benchmark governance, such cases are important because they reveal high-quality evaluators that deserve more systematic reporting despite relatively weak community momentum.

\subsection{Domain-Specific Analysis}

This section provides a domain analysis of 158 benchmarks, offering strategic insights for the design of next-generation evaluation suites and the selection of performance metrics. 

\subsubsection{Mathematics}
In the mathematical domain, a moderate negative correlation ($r = -0.22$) between Impact and Anti-Saturation suggests that high-impact assets tend toward technical obsolescence. The inherent logical consistency of mathematical reasoning likely facilitates rapid pattern internalization by successor models once these assets are extensively benchmarked.

\paragraph{GSM8K \cite{cobbe2021trainingverifierssolvemath}}
BHI metrics characterize GSM8K as a saturated legacy benchmark. Its minimal anti-saturation score ($S_{\mathrm{AS}} = 0.0865$) and low overall $BHI$ ($0.2199$) quantify a terminal loss of diagnostic utility for frontier model architectures. Although $S_{\mathrm{Disc}}$ remains at $0.3920$, the data indicates that GSM8K has transitioned from a research probe into a historical baseline representing past consensus rather than modern technical challenges.

\paragraph{FrontierMath \cite{glazer2025frontiermathbenchmarkevaluatingadvanced}}
FrontierMath exhibits high technical resilience, defined by a robust anti-saturation score ($S_{\mathrm{AS}} = 0.6102$). While its current impact score ($S_{\mathrm{Imp}} = 0.3339$) reflects an early adoption phase, the $BHI$ of $0.5093$ highlights its capacity to resist pattern matching. The indicators suggest that by requiring precise mathematical objects, the benchmark maintains the necessary evaluative headroom for future model scaling.

\subsubsection{Agent}
The agentic and tool-use domain exhibits a healthy positive correlation ($r = 0.30$) between Impact and Anti-Saturation, suggesting that the community consensus is anchored towards benchmarks that maintain high technical difficulty and environmental complexity.

\paragraph{Tau-bench Series \cite{yao2024tau,barres2025tau2}}
The Tau-bench series illustrates how metric profiles reveal the divergent evaluation longevity of sub-scenarios. The Airline subset maintains an $S_{\mathrm{AS}}$ of $0.4798$, indicating robust resistance to templated learning through its multi-constraint game-theoretic designs, such as complex booking and refund policies. In contrast, the Telecom subset, despite its exceptionally high discrimination score ($S_{\mathrm{Disc}} = 0.9005$), exhibits a rapid decay in $S_{\mathrm{AS}}$ to $0.1108$. This sharp contrast suggests that while its logic is highly precise for ranking, procedural workflows are more susceptible to standardization once model architectures master the underlying patterns, leading to a BHI of $0.4921$.

\paragraph{BrowseComp \cite{wei2025browsecompsimplechallengingbenchmark}}
BrowseComp yields an $S_{\mathrm{Disc}}$ of $0.8042$ and a BHI of $0.5845$, validating the efficacy of its reverse-design methodology. The $S_{\mathrm{AS}}$ of $0.3281$ shows that constructing queries from rare facts effectively prevents models from relying on internal knowledge or direct keyword matching. By forcing models toward multi-step, creative search strategies in dynamic web environments, the benchmark maintains strong evaluative vitality and prevents premature saturation of its performance metrics.

\subsubsection{Code}
The paradigm of Code evaluation is shifting, evidenced by the decoupling of $S_{\mathrm{Imp}}$ and $S_{\mathrm{AS}}$, moving from algorithmic puzzles to comprehensive engineering productivity.

\paragraph{HumanEval \cite{chen2021codex} and MBPP \cite{austin2021program}}
BHI diagnostics confirm the diminishing utility of traditional benchmarks. Low $BHI$ scores ($0.2787$ for HumanEval; $0.2371$ for MBPP) and stagnant $S_{\mathrm{AS}}$ ($0.1669$ and $0.1744$, respectively) suggest their focus on function-level algorithms has been neutralized by training data saturation. These metrics indicate that such benchmarks can no longer effectively distinguish genuine reasoning from memorized patterns, as performance gaps narrow below the threshold of effective discrimination.

\paragraph{SWE-bench \cite{jimenez2024swebench}}
SWE-bench evaluates engineering proficiency through real-world issue resolution in large-scale repositories. Its diagnostic profile, characterized by $BHI=0.5293$ and $S_{\mathrm{Disc}}=0.6093$, stems from an execution-based scoring mechanism. Significant $S_{\mathrm{Imp}}$ ($0.7336$) and $S_{\mathrm{AS}}$ ($0.2754$) scores quantify the necessity for a full engineering loop, including fault localization and unit testing. These indicators confirm the benchmark's capacity to target high-level productivity and complex environment navigation.

\subsubsection{Subject}
The Subject domain is the most susceptible to data contamination, with a median $S_{\mathrm{AS}}$ approximately $53.75\%$ lower than the overall median across all evaluation sets.

\paragraph{C-Eval \cite{huang2023ceval}}
C-Eval faces a similar obsolescence crisis (BHI $0.2098$), driven by a critical collapse in its anti-saturation metric ($S_{\mathrm{AS}}$ to $0.0935$). This indicates significant training data leakage. Consequently, C-Eval has lost its stratifying power ($S_{\mathrm{Disc}}$ of $0.2674$), evolving into a memorization check rather than a gauge of generalized subject mastery.

\paragraph{MMLU-Pro \cite{10.5555/3737916.3740934}}
In contrast, MMLU-Pro (BHI $0.3919$) demonstrates the potential for rehabilitating subject-based benchmarks. By increasing reasoning complexity and expanding the option space, it restores discriminatory utility ($S_{\mathrm{Disc}}$ of $0.4572$), significantly exceeding the domain average. Although its $S_{\mathrm{AS}}$ ($0.1727$) suggests contamination remains a challenge, the recovery in discrimination confirms MMLU-Pro’s efficacy in distinguishing capabilities among top-tier models.

\subsubsection{Application}
The Application domain benchmarks evaluate models under real-world deployment conditions, spanning agent-oriented customer service pipelines, financial information retrieval, and high-stakes medical consultation. The domain exhibits a moderate positive correlation ($r = 0.21$) between Impact and Anti-Saturation, suggesting that practical deployment pressures tend to favor benchmarks whose task complexity resists rapid memorization.

\paragraph{GDPval-AA Elo \cite{patwardhan2025gdpvalevaluatingaimodel}}
GDPval-AA Elo achieves a competitive BHI of $0.4225$ with a well-rounded diagnostic profile. Its $S_{\mathrm{Disc}}$ of $0.5962$ demonstrates reliable stratification power across frontier models, while an $S_{\mathrm{Imp}}$ of $0.3409$ reflects meaningful community adoption, confirming that LLM-as-judge frameworks grounded in calibrated preference signals are gaining credibility as an alternative to static answer-key evaluation for open-ended tasks. The $S_{\mathrm{AS}}$ of $0.3362$ indicates moderate evaluative durability; the diversity of judge-model pairings and the use of Elo-based ranking provide a structural buffer against convergence, preserving useful headroom for successive model generations. Taken together, GDPval-AA Elo exemplifies how human-preference-calibrated evaluation can deliver both practical utility and sustained discriminatory power at the frontier.

\paragraph{HealthBench Hard \cite{arora2025healthbenchevaluatinglargelanguage}}
HealthBench Hard stands out within the Application domain for its exceptional anti-saturation properties, recording an $S_{\mathrm{AS}}$ of $0.5924$ and a BHI of $0.4546$. Its scenario-grounded design—requiring expert-validated clinical judgment across low-resource and high-stakes medical contexts—renders surface-level pattern matching ineffective and preserves meaningful performance headroom. The $S_{\mathrm{Disc}}$ of $0.5876$ confirms that the benchmark retains strong differentiating power among frontier systems. Although its $S_{\mathrm{Imp}}$ of $0.1617$ reflects an early adoption phase, the high $S_{\mathrm{AS}}$ indicates that the benchmark's complexity is intrinsic rather than incidental, positioning it as a durable and rigorous instrument for tracking genuine medical reasoning capacity over successive model generations.

\paragraph{HealthBench \cite{arora2025healthbenchevaluatinglargelanguage}}
The base HealthBench benchmark presents a notably weaker profile, with a BHI of $0.3524$ and an $S_{\mathrm{Disc}}$ of $0.4479$, indicating insufficient stratifying power at the current performance frontier. While its $S_{\mathrm{AS}}$ of $0.4016$ remains moderate, the low $S_{\mathrm{Imp}}$ of $0.1983$ signals limited penetration in active leaderboard circuits. As frontier models converge on the primary evaluation scenarios, HealthBench is gradually transitioning toward a legacy role, with its discriminatory utility increasingly superseded by its harder counterpart. These indicators suggest that for frontier-model evaluation purposes, HealthBench Hard has largely displaced the base version as the more informative diagnostic tool.


\section{Compute Resources}
\label{app:compute}

All experiments in this work are lightweight post-hoc meta-analytic computations over publicly reported benchmark scores. We do not train, fine-tune, or run inference with any large language model, and no GPU resources are required. The BHI computation, ablation study, robustness analysis, and sensitivity analysis can be reproduced on a standard CPU workstation or laptop with ordinary memory requirements. The total runtime is dominated by data parsing and repeated recomputation of the BHI pipeline rather than model training or benchmark execution.

\section{Temporal Robustness}
\label{app:temporal_robustness}

\subsection{Updated Data and Evaluation Protocol}

To examine how BHI changes as new evaluation results become available, we extend the original April 2026 submission snapshot through June 2026. The update adds 16 models, increasing the model count from 117 to 133. Candidate benchmarks increase from 416 to 486, while eligible benchmarks increase from 158 to 185, a net increase of 27. We follow the data pipeline in Section~4 and Appendix~E, retaining the same alignment, normalization, and inclusion rules. Metric definitions and hyperparameters remain unchanged. We recompute the complete pipeline, including LOBO capability calibration, component scores, and CRITIC weights.

\subsection{Stability of Component Weights}

CRITIC weights are estimated from each snapshot and may change with the available evidence. Table~\ref{tab:temporal_update_weights} summarizes the update in data coverage and component weights.

\begin{table}[htbp]
\centering
\caption{Data coverage and CRITIC weights in the original submission and the June 2026 update. Weight changes are measured in percentage points (pp).}
\label{tab:temporal_update_weights}
\begin{tabular}{lccc}
\toprule
Indicator & Original submission & June 2026 & Change \\
\midrule
Models & 117 & 133 & +16 \\
Eligible benchmarks & 158 & 185 & +27 \\
$w_{\mathrm{Disc}}$ & 32.63\% & 32.72\% & +0.09 pp \\
$w_{\mathrm{AS}}$ & 35.74\% & 37.45\% & +1.71 pp \\
$w_{\mathrm{Imp}}$ & 31.63\% & 29.83\% & $-1.80$ pp \\
\bottomrule
\end{tabular}
\end{table}

The largest absolute weight change is 1.80 percentage points, suggesting that the overall balance among the three components is largely preserved in this update. These weights are snapshot-specific estimates rather than fixed constants.

\subsection{Stability of Benchmark Rankings and Scores}

Because newly included benchmarks can affect positions in the full ranking, we match benchmarks across the two snapshots and compare rankings recomputed within the common set. We also compare their normalized rank percentiles in the respective full rankings.

Within the common set, Spearman correlation is 0.9848 and Kendall's $\tau_b$ is 0.9159, with a median absolute rank change of 3. The median absolute rank-percentile change is 2.06 percentage points. Among benchmarks originally in the top quartile and top half of the common set, 89.47\% and 96.05\%, respectively, remain in those groups. Top-10 and Top-20 overlap between the full rankings is 7/10 and 14/20.

BHI scores also remain highly consistent across the matched benchmarks, with a Pearson correlation of 0.9862. The mean and median absolute BHI changes are 0.0122 and 0.0064 on the $[0,1]$ scale, and 95.36\% of changes are below 0.05. These results suggest that ranking consistency is accompanied by small absolute score changes.

At the component level, Anti-Saturation is the most stable, while Discrimination is more responsive to additional model scores. Impact may also change as evaluation coverage and LOBO capability estimates are updated. These results reflect both robustness, with the global weighting and ranking structure remaining highly consistent, and dynamism, as new evidence can appropriately change eligibility, weights, scores, and ranks.

\newpage

{\scriptsize
\setlength{\tabcolsep}{1.5pt}
\begin{longtable}{@{}p{0.45cm} p{3.3cm} >{\centering\arraybackslash}p{2.8cm} >{\centering\arraybackslash}p{3.2cm} >{\centering\arraybackslash}p{3.2cm} >{\raggedleft\arraybackslash}p{0.85cm}@{}}
    \caption{Detailed BHI Metric Results and Sub-metric Results}
    \label{tab:appendix-bhi_full_decomposition_case_study} \\
    \toprule
    Rank & Benchmark & \textbf{$S_{\mathrm{Disc}}$} ($RCV$, $EDR$) & \textbf{$S_{\mathrm{AS}}$} ($S_{\mathrm{Sta}}$, $S_{\mathrm{Dyn}}$) & \textbf{$S_{\mathrm{Imp}}$} ($N_{\mathrm{Usage}}$) & \textbf{BHI} \\
    \midrule
    \endfirsthead

    \multicolumn{6}{c}{\small\tablename\ \thetable{} -- continued from previous page} \\
    \toprule
    Rank & Benchmark & \textbf{$S_{\mathrm{Disc}}$} ($RCV$, $EDR$) & \textbf{$S_{\mathrm{AS}}$} ($S_{\mathrm{Sta}}$, $S_{\mathrm{Dyn}}$) & \textbf{$S_{\mathrm{Imp}}$} ($N_{\mathrm{Usage}}$) & \textbf{BHI} \\
    \midrule
    \endhead

    \midrule
    \multicolumn{6}{r}{\small Continued on next page} \\
    \endfoot

    \bottomrule
    \endlastfoot

    1 & Humanity's Last Exam \protect\cite{phan2025humanitysexam} & \textbf{0.6608} (0.2954, 0.9131) & \textbf{0.6299} (0.6211, 0.6653) & \textbf{0.7710} (56) & \textbf{0.6846} \\
    2 & SWE-Lancer(no limited) \protect\cite{miserendino2025swelancerfrontierllmsearn} & \textbf{0.9979} (0.6396, 1.0000) & \textbf{0.6102} (0.6129, 0.5992) & \textbf{0.1757} (3) & \textbf{0.5992} \\
    3 & SimpleQA \protect\cite{wei2024measuringshortformfactualitylarge} & \textbf{0.8130} (0.4599, 0.9347) & \textbf{0.5043} (0.5011, 0.5169) & \textbf{0.4435} (27) & \textbf{0.5857} \\
    4 & BrowseComp \protect\cite{wei2025browsecompsimplechallengingbenchmark} & \textbf{0.8042} (0.4490, 0.9365) & \textbf{0.3281} (0.3200, 0.3607) & \textbf{0.6475} (36) & \textbf{0.5845} \\
    5 & ARC-AGI-2 \protect\cite{chollet2026arcagi2newchallengefrontier} & \textbf{0.8531} (0.4871, 0.9714) & \textbf{0.4411} (0.4487, 0.4104) & \textbf{0.4513} (15) & \textbf{0.5787} \\
    6 & LiveCodeBench-V6 \protect\cite{jain2024livecodebench} & \textbf{0.8266} (0.4733, 0.9393) & \textbf{0.2786} (0.2680, 0.3213) & \textbf{0.6593} (56) & \textbf{0.5779} \\
    7 & ZeroBench \protect\cite{roberts2025zerobenchimpossiblevisualbenchmark} & \textbf{0.5205} (0.1925, 0.8000) & \textbf{0.8687} (0.8594, 0.9059) & \textbf{0.2498} (6) & \textbf{0.5593} \\
    8 & Graphwalks bfs>128k \protect\cite{openai2024graphwalks} & \textbf{0.8206} (0.4348, 1.0000) & \textbf{0.5978} (0.5656, 0.7268) & \textbf{0.2241} (4) & \textbf{0.5523} \\
    9 & SWE-Bench \protect\cite{jimenez2024swebench} & \textbf{0.6093} (0.2615, 0.8640) & \textbf{0.2754} (0.2712, 0.2923) & \textbf{0.7336} (57) & \textbf{0.5293} \\
    10 & Graphwalks parents>128k \protect\cite{openai2024graphwalks} & \textbf{0.9415} (0.5745, 1.0000) & \textbf{0.4097} (0.3758, 0.5451) & \textbf{0.2241} (4) & \textbf{0.5245} \\
    11 & Terminal-Bench-V2 \protect\cite{merrill2026terminalbenchbenchmarkingagentshard} & \textbf{0.5848} (0.2174, 0.8945) & \textbf{0.4077} (0.4060, 0.4145) & \textbf{0.5880} (27) & \textbf{0.5225} \\
    12 & MRCR-V2 1M \protect\cite{openai2024mrcr} & \textbf{0.5364} (0.1760, 0.8666) & \textbf{0.7508} (0.7525, 0.7441) & \textbf{0.2409} (6) & \textbf{0.5195} \\
    13 & GPQA-Diamond \protect\cite{rein2024gpqa} & \textbf{0.5985} (0.2429, 0.8759) & \textbf{0.1638} (0.1588, 0.1839) & \textbf{0.8175} (74) & \textbf{0.5124} \\
    14 & FrontierMath \protect\cite{glazer2025frontiermathbenchmarkevaluatingadvanced} & \textbf{0.5686} (0.2162, 0.8611) & \textbf{0.6102} (0.6040, 0.6352) & \textbf{0.3339} (9) & \textbf{0.5093} \\
    15 & Toolathlon \protect\cite{li2025toolathlon} & \textbf{0.5162} (0.1519, 0.8681) & \textbf{0.5414} (0.5369, 0.5595) & \textbf{0.4258} (14) & \textbf{0.4966} \\
    16 & AIME 2025 \protect\cite{aime25} & \textbf{0.6923} (0.3445, 0.8889) & \textbf{0.1154} (0.1122, 0.1284) & \textbf{0.7194} (61) & \textbf{0.4947} \\
    17 & VitaBench \protect\cite{he2026vitabench} & \textbf{0.6375} (0.2232, 1.0000) & \textbf{0.5846} (0.5864, 0.5773) & \textbf{0.2445} (5) & \textbf{0.4943} \\
    18 & AIME 2024 \protect\cite{jia2024aime} & \textbf{0.8522} (0.5187, 0.9090) & \textbf{0.2093} (0.2064, 0.2210) & \textbf{0.4450} (33) & \textbf{0.4936} \\
    19 & Terminal-Bench \protect\cite{merrill2026terminalbenchbenchmarkingagentshard} & \textbf{0.5308} (0.1596, 0.8857) & \textbf{0.6012} (0.5989, 0.6104) & \textbf{0.3336} (15) & \textbf{0.4936} \\
    20 & Tau2-bench-Telecom \protect\cite{barres2025tau2} & \textbf{0.9005} (0.6419, 0.7800) & \textbf{0.1108} (0.1058, 0.1309) & \textbf{0.5016} (25) & \textbf{0.4921} \\
    21 & OJBench \protect\cite{wang2025ojbenchcompetitionlevelcode} & \textbf{0.6453} (0.2669, 0.9333) & \textbf{0.5404} (0.5305, 0.5803) & \textbf{0.2620} (7) & \textbf{0.4866} \\
    22 & ARC-AGI \protect\cite{chollet2019measureintelligence} & \textbf{0.8734} (0.5433, 0.9090) & \textbf{0.2249} (0.2087, 0.2898) & \textbf{0.3554} (12) & \textbf{0.4778} \\
    23 & Multi-SWE-Bench \protect\cite{zan2025multiswebench} & \textbf{0.5767} (0.2088, 0.8928) & \textbf{0.5551} (0.5590, 0.5395) & \textbf{0.2740} (8) & \textbf{0.4732} \\
    24 & HMMT NOV 2025 \protect\cite{balunovic2025matharena} & \textbf{0.8329} (0.5075, 0.8876) & \textbf{0.1281} (0.1223, 0.1513) & \textbf{0.4879} (27) & \textbf{0.4719} \\
    25 & ZeroBench(sub) \protect\cite{roberts2025zerobenchimpossiblevisualbenchmark} & \textbf{0.6146} (0.1968, 1.0000) & \textbf{0.5882} (0.5863, 0.5959) & \textbf{0.1790} (3) & \textbf{0.4674} \\
    26 & SciCode \protect\cite{tian2024scicode} & \textbf{0.5925} (0.1712, 1.0000) & \textbf{0.5201} (0.5175, 0.5304) & \textbf{0.2767} (7) & \textbf{0.4667} \\
    27 & Tau1-bench-Airline \protect\cite{yao2024tau} & \textbf{0.5869} (0.2343, 0.8666) & \textbf{0.4798} (0.4783, 0.4857) & \textbf{0.3053} (15) & \textbf{0.4595} \\
    28 & VisuLogic \protect\cite{xu2026visulogic} & \textbf{0.5752} (0.1512, 1.0000) & \textbf{0.6099} (0.6064, 0.6242) & \textbf{0.1674} (3) & \textbf{0.4586} \\
    29 & BrowseComp-ZH \protect\cite{zhou2025browsecomp} & \textbf{0.6912} (0.3248, 0.9242) & \textbf{0.3380} (0.3431, 0.3176) & \textbf{0.3492} (12) & \textbf{0.4568} \\
    30 & CyberGym \protect\cite{wang2026cybergym} & \textbf{0.7188} (0.3519, 0.9333) & \textbf{0.3647} (0.3573, 0.3947) & \textbf{0.2889} (6) & \textbf{0.4563} \\
    31 & Tau2-bench-Airline \protect\cite{yao2024tau} & \textbf{0.6896} (0.3230, 0.9242) & \textbf{0.3813} (0.3805, 0.3847) & \textbf{0.2978} (12) & \textbf{0.4555} \\
    32 & Vending-Bench 2(no limited) \protect\cite{backlund2025vending} & \textbf{0.6677} (0.3103, 0.9000) & \textbf{0.4391} (0.4446, 0.4170) & \textbf{0.2526} (5) & \textbf{0.4547} \\
    33 & HealthBench Hard \protect\cite{arora2025healthbenchevaluatinglargelanguage} & \textbf{0.5876} (0.1656, 1.0000) & \textbf{0.5924} (0.5801, 0.6415) & \textbf{0.1617} (3) & \textbf{0.4546} \\
    34 & MultiChallenge \protect\cite{deshpande-etal-2025-multichallenge} & \textbf{0.5893} (0.2470, 0.8476) & \textbf{0.4087} (0.4065, 0.4178) & \textbf{0.3431} (15) & \textbf{0.4469} \\
    35 & Aider Polyglot \protect\cite{gauthier2024aiderpolyglot} & \textbf{0.6699} (0.2930, 0.9380) & \textbf{0.2924} (0.2871, 0.3137) & \textbf{0.3818} (21) & \textbf{0.4438} \\
    36 & PolyMATH \protect\cite{gupta2024polymath} & \textbf{0.7387} (0.3401, 1.0000) & \textbf{0.3495} (0.3399, 0.3879) & \textbf{0.2329} (7) & \textbf{0.4396} \\
    37 & Codeforces-Rating(no limited) \protect\cite{quan2025codeelo} & \textbf{0.6104} (0.2615, 0.8666) & \textbf{0.3695} (0.3584, 0.4138) & \textbf{0.3323} (12) & \textbf{0.4363} \\
    38 & MRCR-V2 128K \protect\cite{openai2024mrcr} & \textbf{0.7021} (0.3414, 0.9166) & \textbf{0.3174} (0.3132, 0.3343) & \textbf{0.2957} (9) & \textbf{0.4360} \\
    39 & Beyond AIME \protect\cite{bytedance_seed_2025_beyondaime} & \textbf{0.7706} (0.3770, 1.0000) & \textbf{0.3218} (0.3171, 0.3405) & \textbf{0.2056} (5) & \textbf{0.4315} \\
    40 & SWE-Bench Multilingual \protect\cite{jimenez2024swebench} & \textbf{0.5292} (0.2000, 0.8047) & \textbf{0.2964} (0.2961, 0.2975) & \textbf{0.4748} (21) & \textbf{0.4288} \\
    41 & DynaMath \protect\cite{zou2025dynamath} & \textbf{0.7188} (0.3731, 0.8928) & \textbf{0.3033} (0.2969, 0.3292) & \textbf{0.2687} (8) & \textbf{0.4279} \\
    42 & SuperGPQA \protect\cite{du2025supergpqa} & \textbf{0.5686} (0.1915, 0.9083) & \textbf{0.3680} (0.3638, 0.3848) & \textbf{0.3369} (16) & \textbf{0.4236} \\
    43 & GDPval-AA Elo(no limited) \protect\cite{patwardhan2025gdpvalevaluatingaimodel} & \textbf{0.5962} (0.2219, 0.9111) & \textbf{0.3362} (0.3479, 0.2897) & \textbf{0.3409} (9) & \textbf{0.4225} \\
    44 & SWE-Bench Pro \protect\cite{deng2025swebenchproaiagents} & \textbf{0.3786} (0.0732, 0.7142) & \textbf{0.4459} (0.4448, 0.4504) & \textbf{0.4395} (14) & \textbf{0.4219} \\
    45 & BabyVision \protect\cite{chen2026babyvisionvisualreasoninglanguage} & \textbf{0.6228} (0.2061, 1.0000) & \textbf{0.3949} (0.4436, 0.1998) & \textbf{0.2364} (4) & \textbf{0.4191} \\
    46 & MCP Atlas \protect\cite{bandi2026mcpatlaslargescalebenchmarktooluse} & \textbf{0.4956} (0.1510, 0.8241) & \textbf{0.3219} (0.3204, 0.3281) & \textbf{0.4434} (14) & \textbf{0.4170} \\
    47 & Seal-0 \protect\cite{pham2026sealqa} & \textbf{0.5170} (0.0839, 1.0000) & \textbf{0.4816} (0.4782, 0.4952) & \textbf{0.2254} (4) & \textbf{0.4121} \\
    48 & MRCR 128K \protect\cite{openai2024mrcr} & \textbf{0.7822} (0.3903, 1.0000) & \textbf{0.3076} (0.2992, 0.3413) & \textbf{0.1481} (3) & \textbf{0.4120} \\
    49 & CharXiv-Reasoning \protect\cite{wang2024charxiv} & \textbf{0.5575} (0.2536, 0.7647) & \textbf{0.2287} (0.2217, 0.2565) & \textbf{0.4337} (18) & \textbf{0.4008} \\
    50 & MCP-Mark \protect\cite{wu2026mcpmark} & \textbf{0.4452} (0.0880, 0.8333) & \textbf{0.4927} (0.4860, 0.5194) & \textbf{0.2424} (4) & \textbf{0.3980} \\
    51 & EMMA \protect\cite{hao2025can} & \textbf{0.6638} (0.2536, 1.0000) & \textbf{0.3613} (0.3564, 0.3807) & \textbf{0.1629} (3) & \textbf{0.3972} \\
    52 & MMMU-Pro \protect\cite{yue-etal-2025-mmmu} & \textbf{0.4774} (0.1850, 0.7190) & \textbf{0.2293} (0.2262, 0.2417) & \textbf{0.4994} (21) & \textbf{0.3957} \\
    53 & GPQA \protect\cite{rein2024gpqa} & \textbf{0.6114} (0.2339, 0.9215) & \textbf{0.1979} (0.1916, 0.2234) & \textbf{0.3922} (20) & \textbf{0.3943} \\
    54 & Graphwalks bfs <128k \protect\cite{openai2024graphwalks} & \textbf{0.7355} (0.3712, 0.9333) & \textbf{0.2364} (0.2303, 0.2609) & \textbf{0.2192} (6) & \textbf{0.3938} \\
    55 & MATH-Vision \protect\cite{wang2024measuring} & \textbf{0.6824} (0.3320, 0.8909) & \textbf{0.1860} (0.1767, 0.2229) & \textbf{0.3308} (11) & \textbf{0.3938} \\
    56 & OSWorld \protect\cite{xie2024osworld} & \textbf{0.4907} (0.1579, 0.8000) & \textbf{0.3033} (0.2990, 0.3205) & \textbf{0.3958} (11) & \textbf{0.3937} \\
    57 & MMLU-Pro \protect\cite{10.5555/3737916.3740934} & \textbf{0.4572} (0.1443, 0.7520) & \textbf{0.1727} (0.1704, 0.1822) & \textbf{0.5724} (47) & \textbf{0.3919} \\
    58 & C-SimpleQA \protect\cite{he-etal-2025-chinese} & \textbf{0.5926} (0.2251, 0.8970) & \textbf{0.2297} (0.2237, 0.2536) & \textbf{0.3585} (17) & \textbf{0.3888} \\
    59 & FinSearchComp-T3 \protect\cite{hu2026finsearchcomp} & \textbf{0.5856} (0.1632, 1.0000) & \textbf{0.3907} (0.4054, 0.3317) & \textbf{0.1819} (3) & \textbf{0.3882} \\
    60 & NL2Repo \protect\cite{ding2025nl2repo} & \textbf{0.3269} (0.0384, 0.6666) & \textbf{0.6099} (0.6000, 0.6495) & \textbf{0.1955} (3) & \textbf{0.3865} \\
    61 & OlympiadBench \protect\cite{he-etal-2024-olympiadbench} & \textbf{0.6728} (0.2639, 1.0000) & \textbf{0.3564} (0.3657, 0.3191) & \textbf{0.1142} (3) & \textbf{0.3830} \\
    62 & Tau2-bench-Retail \protect\cite{barres2025tau2} & \textbf{0.5801} (0.2217, 0.8758) & \textbf{0.1726} (0.1706, 0.1808) & \textbf{0.4151} (18) & \textbf{0.3823} \\
    63 & MMMU \protect\cite{Yue_2024_CVPR} & \textbf{0.4509} (0.1253, 0.7746) & \textbf{0.2074} (0.2057, 0.2142) & \textbf{0.5034} (33) & \textbf{0.3805} \\
    64 & MATH \protect\cite{hendrycks2021measuring} & \textbf{0.6585} (0.2759, 0.9454) & \textbf{0.2604} (0.2535, 0.2878) & \textbf{0.2091} (11) & \textbf{0.3741} \\
    65 & LiveCodeBench-Pro(no limited) \protect\cite{zheng2025livecodebench} & \textbf{0.5545} (0.1795, 0.9000) & \textbf{0.3118} (0.3190, 0.2831) & \textbf{0.2503} (5) & \textbf{0.3715} \\
    66 & ScreenSpot-Pro \protect\cite{li2025screenspotpro} & \textbf{0.5726} (0.1829, 0.9333) & \textbf{0.2668} (0.2587, 0.2992) & \textbf{0.2814} (6) & \textbf{0.3712} \\
    67 & FACTS Grounding \protect\cite{kaggle-FACTS-leaderboard} & \textbf{0.6396} (0.2256, 1.0000) & \textbf{0.2664} (0.2647, 0.2732) & \textbf{0.1965} (4) & \textbf{0.3660} \\
    68 & ERQA \protect\cite{geminiroboticsteam2025geminiroboticsbringingai} & \textbf{0.4521} (0.1582, 0.7142) & \textbf{0.3576} (0.3550, 0.3680) & \textbf{0.2798} (7) & \textbf{0.3638} \\
    69 & COLLIE \protect\cite{yao2024collie} & \textbf{0.6588} (0.3224, 0.8571) & \textbf{0.2003} (0.1925, 0.2317) & \textbf{0.2387} (8) & \textbf{0.3620} \\
    70 & Tau1-bench-Retail \protect\cite{yao2024tau} & \textbf{0.5154} (0.1766, 0.8190) & \textbf{0.2719} (0.2705, 0.2774) & \textbf{0.3050} (15) & \textbf{0.3618} \\
    71 & LiveCodeBench-V5 \protect\cite{jain2024livecodebench} & \textbf{0.5259} (0.1523, 0.8888) & \textbf{0.3198} (0.3183, 0.3256) & \textbf{0.2364} (9) & \textbf{0.3606} \\
    72 & BFCL-v3 \protect\cite{patil2025the} & \textbf{0.4971} (0.1886, 0.7555) & \textbf{0.3163} (0.3183, 0.3083) & \textbf{0.2524} (10) & \textbf{0.3551} \\
    73 & Arena-Hard \protect\cite{arenahard2024} & \textbf{0.6220} (0.2790, 0.8589) & \textbf{0.1716} (0.1681, 0.1857) & \textbf{0.2793} (13) & \textbf{0.3526} \\
    74 & HealthBench \protect\cite{arora2025healthbenchevaluatinglargelanguage} & \textbf{0.4479} (0.0911, 0.8333) & \textbf{0.4016} (0.3990, 0.4121) & \textbf{0.1983} (4) & \textbf{0.3524} \\
    75 & Finance Agent v1.1 \protect\cite{bigeard2025financeagentbenchmarkbenchmarking} & \textbf{0.3638} (0.0635, 0.7000) & \textbf{0.4086} (0.3902, 0.4826) & \textbf{0.2754} (5) & \textbf{0.3518} \\
    76 & MathVista \protect\cite{lu2024mathvista} & \textbf{0.5354} (0.1802, 0.8562) & \textbf{0.1698} (0.1663, 0.1837) & \textbf{0.3556} (18) & \textbf{0.3478} \\
    77 & IMO-AnswerBench \protect\cite{luong-etal-2025-towards} & \textbf{0.4576} (0.1154, 0.8083) & \textbf{0.1605} (0.1600, 0.1626) & \textbf{0.4359} (16) & \textbf{0.3446} \\
    78 & ZebraLogic \protect\cite{pmlr-v267-lin25i} & \textbf{0.6394} (0.2728, 0.9090) & \textbf{0.1454} (0.1447, 0.1483) & \textbf{0.2644} (11) & \textbf{0.3442} \\
    79 & WeMath \protect\cite{qiao-etal-2025-math} & \textbf{0.6394} (0.2776, 0.9000) & \textbf{0.1941} (0.1870, 0.2224) & \textbf{0.2094} (5) & \textbf{0.3442} \\
    80 & Tau2-bench \protect\cite{barres2025tau2} & \textbf{0.4920} (0.1495, 0.8190) & \textbf{0.1788} (0.1815, 0.1683) & \textbf{0.3728} (15) & \textbf{0.3423} \\
    81 & MathVerse \protect\cite{10.1007/978-3-031-73242-3_10} & \textbf{0.6092} (0.1904, 1.0000) & \textbf{0.2413} (0.2388, 0.2514) & \textbf{0.1612} (4) & \textbf{0.3360} \\
    82 & BFCL-v4 \protect\cite{pmlr-v267-patil25a} & \textbf{0.4838} (0.1326, 0.8333) & \textbf{0.3026} (0.3161, 0.2483) & \textbf{0.2195} (4) & \textbf{0.3354} \\
    83 & Graphwalks parents <128k \protect\cite{openai2024graphwalks} & \textbf{0.6643} (0.3063, 0.9000) & \textbf{0.1287} (0.1204, 0.1617) & \textbf{0.2245} (5) & \textbf{0.3337} \\
    84 & MRCR 1M \protect\cite{openai2024mrcr} & \textbf{0.5458} (0.2042, 0.8333) & \textbf{0.2486} (0.2108, 0.3995) & \textbf{0.2024} (4) & \textbf{0.3309} \\
    85 & SimpleVQA \protect\cite{cheng2025simplevqa} & \textbf{0.3748} (0.0688, 0.7142) & \textbf{0.3151} (0.3147, 0.3166) & \textbf{0.3027} (8) & \textbf{0.3307} \\
    86 & LogicVista \protect\cite{xiao2024logicvista} & \textbf{0.5584} (0.2188, 0.8333) & \textbf{0.2495} (0.2422, 0.2788) & \textbf{0.1730} (4) & \textbf{0.3261} \\
    87 & MMStar \protect\cite{10.5555/3737916.3738766} & \textbf{0.4733} (0.1189, 0.8363) & \textbf{0.1933} (0.1910, 0.2027) & \textbf{0.3239} (11) & \textbf{0.3260} \\
    88 & LongBench-v2 \protect\cite{bai-etal-2025-longbench} & \textbf{0.2882} (0.0433, 0.5714) & \textbf{0.3808} (0.3807, 0.3812) & \textbf{0.3003} (8) & \textbf{0.3251} \\
    89 & MultiPL-E \protect\cite{10103177} & \textbf{0.5541} (0.2014, 0.8571) & \textbf{0.2408} (0.2423, 0.2349) & \textbf{0.1748} (7) & \textbf{0.3221} \\
    90 & IFBench \protect\cite{pyatkin2026generalizing} & \textbf{0.4609} (0.0714, 0.9000) & \textbf{0.2745} (0.2747, 0.2739) & \textbf{0.2300} (5) & \textbf{0.3213} \\
    91 & Hallusion Bench \protect\cite{Guan_2024_CVPR} & \textbf{0.3716} (0.0652, 0.7142) & \textbf{0.3377} (0.3365, 0.3429) & \textbf{0.2477} (7) & \textbf{0.3203} \\
    92 & WideSearch \protect\cite{wong2026widesearch} & \textbf{0.3969} (0.0943, 0.7142) & \textbf{0.2617} (0.2637, 0.2541) & \textbf{0.3031} (7) & \textbf{0.3189} \\
    93 & MMVU \protect\cite{zhao2025mmvu} & \textbf{0.4591} (0.0693, 0.9000) & \textbf{0.2367} (0.2388, 0.2280) & \textbf{0.2457} (5) & \textbf{0.3121} \\
    94 & Vibe-Eval \protect\cite{padlewski2024vibeevalhardevaluationsuite} & \textbf{0.4716} (0.1186, 0.8333) & \textbf{0.2712} (0.2695, 0.2779) & \textbf{0.1898} (4) & \textbf{0.3108} \\
    95 & Blink \protect\cite{10.1007/978-3-031-73337-6_9} & \textbf{0.4423} (0.1021, 0.8000) & \textbf{0.2497} (0.2492, 0.2517) & \textbf{0.2309} (5) & \textbf{0.3066} \\
    96 & MBPP+ \protect\cite{mbppplus10.1145/3680410} & \textbf{0.4738} (0.1065, 0.8611) & \textbf{0.2218} (0.2197, 0.2300) & \textbf{0.2271} (9) & \textbf{0.3057} \\
    97 & DeepSearchQA \protect\cite{gupta2026deepsearchqabridgingcomprehensivenessgap} & \textbf{0.4813} (0.0950, 0.9000) & \textbf{0.1897} (0.2031, 0.1360) & \textbf{0.2542} (5) & \textbf{0.3052} \\
    98 & GDPval \protect\cite{patwardhan2025gdpvalevaluatingaimodel} & \textbf{0.4229} (0.1145, 0.7333) & \textbf{0.1987} (0.2011, 0.1892) & \textbf{0.2981} (6) & \textbf{0.3033} \\
    99 & RefSpatialBench \protect\cite{zhou2026roborefer} & \textbf{0.4134} (0.1383, 0.6666) & \textbf{0.2974} (0.3118, 0.2395) & \textbf{0.1922} (3) & \textbf{0.3020} \\
    100 & AA-LCR \protect\cite{artificialanalysis2025lcr} & \textbf{0.3567} (0.0728, 0.6666) & \textbf{0.3369} (0.3378, 0.3335) & \textbf{0.2056} (4) & \textbf{0.3018} \\
    101 & Creative Writing v3(no limited) \protect\cite{creative-writing-bench-v3} & \textbf{0.4626} (0.1255, 0.8000) & \textbf{0.2860} (0.2942, 0.2533) & \textbf{0.1463} (6) & \textbf{0.2994} \\
    102 & MMMLU \protect\cite{hendrycks2021measuring} & \textbf{0.3628} (0.1170, 0.5955) & \textbf{0.1138} (0.1128, 0.1179) & \textbf{0.4434} (17) & \textbf{0.2993} \\
    103 & AGI-Eval \protect\cite{zhong-etal-2024-agieval} & \textbf{0.5973} (0.1767, 1.0000) & \textbf{0.1846} (0.1827, 0.1922) & \textbf{0.1208} (3) & \textbf{0.2991} \\
    104 & IFEval \protect\cite{zhou2023instructionfollowingevaluationlargelanguage} & \textbf{0.4070} (0.0968, 0.7318) & \textbf{0.1078} (0.1068, 0.1117) & \textbf{0.4007} (24) & \textbf{0.2981} \\
    105 & HumanEval+ \protect\cite{HumanEval+10.5555/3666122.3667065} & \textbf{0.5750} (0.2007, 0.9047) & \textbf{0.1282} (0.1235, 0.1471) & \textbf{0.2007} (7) & \textbf{0.2969} \\
    106 & DROP \protect\cite{dua-etal-2019-drop} & \textbf{0.4988} (0.1433, 0.8461) & \textbf{0.1291} (0.1250, 0.1452) & \textbf{0.2692} (13) & \textbf{0.2940} \\
    107 & MultiIF \protect\cite{he2024multiifbenchmarkingllmsmultiturn} & \textbf{0.4401} (0.1257, 0.7500) & \textbf{0.2128} (0.2104, 0.2224) & \textbf{0.2209} (8) & \textbf{0.2895} \\
    108 & CNMO 2024 \protect\cite{lee2025qwen3cnmo} & \textbf{0.5334} (0.1029, 1.0000) & \textbf{0.1881} (0.1879, 0.1886) & \textbf{0.1526} (4) & \textbf{0.2895} \\
    109 & MGSM \protect\cite{shi2023language} & \textbf{0.5786} (0.1551, 1.0000) & \textbf{0.1785} (0.1785, 0.1784) & \textbf{0.1154} (3) & \textbf{0.2891} \\
    110 & Claw-Eval \protect\cite{ye2026clawevaltrustworthyevaluationautonomous} & \textbf{0.3857} (0.1064, 0.6666) & \textbf{0.2737} (0.3338, 0.0334) & \textbf{0.2035} (3) & \textbf{0.2881} \\
    111 & GAIA(text only) \protect\cite{mialon2023gaia} & \textbf{0.5253} (0.0935, 1.0000) & \textbf{0.1605} (0.1639, 0.1468) & \textbf{0.1625} (3) & \textbf{0.2801} \\
    112 & CLUEWSC \protect\cite{xu-etal-2020-clue} & \textbf{0.5417} (0.1623, 0.9047) & \textbf{0.1173} (0.1141, 0.1298) & \textbf{0.1902} (7) & \textbf{0.2788} \\
    113 & HumanEval \protect\cite{chen2021codex} & \textbf{0.5520} (0.1243, 1.0000) & \textbf{0.1669} (0.1710, 0.1508) & \textbf{0.1232} (4) & \textbf{0.2787} \\
    114 & LiveBench \protect\cite{livebench} & \textbf{0.3968} (0.1540, 0.6000) & \textbf{0.2590} (0.2601, 0.2545) & \textbf{0.1741} (6) & \textbf{0.2771} \\
    115 & HMMT Feb 26 \protect\cite{balunovic2025matharena} & \textbf{0.4888} (0.1036, 0.9000) & \textbf{0.0919} (0.0919, 0.0919) & \textbf{0.2679} (5) & \textbf{0.2770} \\
    116 & WritingBench \protect\cite{wu2025writingbench} & \textbf{0.5119} (0.1129, 0.9333) & \textbf{0.1358} (0.1338, 0.1440) & \textbf{0.1864} (6) & \textbf{0.2745} \\
    117 & Video-MMMU \protect\cite{hu2025videommmuevaluatingknowledgeacquisition} & \textbf{0.3136} (0.0579, 0.6000) & \textbf{0.1543} (0.1541, 0.1551) & \textbf{0.3583} (11) & \textbf{0.2708} \\
    118 & INCLUDE \protect\cite{10.1145/3394171.3413528} & \textbf{0.4249} (0.0819, 0.8000) & \textbf{0.1732} (0.1706, 0.1833) & \textbf{0.2220} (6) & \textbf{0.2707} \\
    119 & SysBench \protect\cite{qin2025sysbench} & \textbf{0.3928} (0.0971, 0.7000) & \textbf{0.2216} (0.2247, 0.2095) & \textbf{0.1832} (5) & \textbf{0.2653} \\
    120 & MotionBench \protect\cite{hong2025motionbench} & \textbf{0.3269} (0.0384, 0.6666) & \textbf{0.2733} (0.2767, 0.2597) & \textbf{0.1900} (3) & \textbf{0.2644} \\
    121 & MMLU-ProX \protect\cite{xuan-etal-2025-mmluprox} & \textbf{0.3965} (0.0839, 0.7333) & \textbf{0.1768} (0.1741, 0.1877) & \textbf{0.2220} (6) & \textbf{0.2628} \\
    122 & HMMT Feb 2025 \protect\cite{balunovic2025matharena} & \textbf{0.3955} (0.0940, 0.7121) & \textbf{0.0401} (0.0407, 0.0379) & \textbf{0.3764} (12) & \textbf{0.2625} \\
    123 & RealWorldQA \protect\cite{visheratin2024realworldqa} & \textbf{0.3738} (0.0752, 0.7000) & \textbf{0.1666} (0.1656, 0.1705) & \textbf{0.2379} (5) & \textbf{0.2567} \\
    124 & FRAMES \protect\cite{krishna-etal-2025-fact-frames} & \textbf{0.3702} (0.0709, 0.7000) & \textbf{0.1720} (0.1715, 0.1739) & \textbf{0.2232} (5) & \textbf{0.2528} \\
    125 & Video-MME \protect\cite{fu2025video} & \textbf{0.3374} (0.1003, 0.5714) & \textbf{0.1402} (0.1403, 0.1397) & \textbf{0.2870} (7) & \textbf{0.2510} \\
    126 & MMLU \protect\cite{hendrycks2021measuring} & \textbf{0.3215} (0.0620, 0.6095) & \textbf{0.1193} (0.1188, 0.1213) & \textbf{0.3248} (21) & \textbf{0.2502} \\
    127 & MMBench(v1.1) \protect\cite{10.1007/978-3-031-72658-3_13} & \textbf{0.4985} (0.0626, 1.0000) & \textbf{0.0869} (0.0852, 0.0936) & \textbf{0.1787} (4) & \textbf{0.2502} \\
    128 & PIQA \protect\cite{kastryulin2022pytorchimagequalitymetrics} & \textbf{0.4961} (0.1121, 0.9000) & \textbf{0.1036} (0.1024, 0.1085) & \textbf{0.1409} (5) & \textbf{0.2435} \\
    129 & MBPP \protect\cite{austin2021program} & \textbf{0.4225} (0.0618, 0.8333) & \textbf{0.1744} (0.1749, 0.1726) & \textbf{0.1168} (4) & \textbf{0.2371} \\
    130 & LVBench \protect\cite{Wang_2025_ICCV} & \textbf{0.2400} (0.0250, 0.5000) & \textbf{0.2437} (0.2464, 0.2330) & \textbf{0.2226} (4) & \textbf{0.2358} \\
    131 & LongVideoBench \protect\cite{wu2024longvideobench} & \textbf{0.3137} (0.0231, 0.6666) & \textbf{0.2038} (0.2066, 0.1928) & \textbf{0.1901} (3) & \textbf{0.2353} \\
    132 & AI2D \protect\cite{kembhavi2016diagram} & \textbf{0.4149} (0.0705, 0.8000) & \textbf{0.0681} (0.0680, 0.0685) & \textbf{0.2341} (5) & \textbf{0.2337} \\
    133 & AceBench \protect\cite{chen-etal-2025-acebench} & \textbf{0.3871} (0.1079, 0.6666) & \textbf{0.1703} (0.1698, 0.1723) & \textbf{0.1458} (3) & \textbf{0.2333} \\
    134 & CharXiv-Descriptive \protect\cite{wang2024charxiv} & \textbf{0.4090} (0.0635, 0.8000) & \textbf{0.1031} (0.1022, 0.1068) & \textbf{0.1989} (5) & \textbf{0.2332} \\
    135 & WinoGrande \protect\cite{10.1145/3474381} & \textbf{0.4046} (0.1281, 0.6666) & \textbf{0.1515} (0.1489, 0.1616) & \textbf{0.1316} (4) & \textbf{0.2278} \\
    136 & OCRBench \protect\cite{fu2026ocrbench} & \textbf{0.3649} (0.0649, 0.7000) & \textbf{0.0967} (0.0968, 0.0965) & \textbf{0.2264} (5) & \textbf{0.2252} \\
    137 & MATH-500 \protect\cite{lightman2024lets} & \textbf{0.3279} (0.1237, 0.5054) & \textbf{0.0717} (0.0746, 0.0603) & \textbf{0.2819} (14) & \textbf{0.2218} \\
    138 & CRUX-O \protect\cite{pmlr-v235-gu24c} & \textbf{0.3159} (0.0257, 0.6666) & \textbf{0.2307} (0.2301, 0.2333) & \textbf{0.1142} (3) & \textbf{0.2217} \\
    139 & MMLU-Redux \protect\cite{gema-etal-2025-done} & \textbf{0.2812} (0.0651, 0.5142) & \textbf{0.0728} (0.0714, 0.0786) & \textbf{0.3230} (15) & \textbf{0.2199} \\
    140 & GSM8K \protect\cite{cobbe2021trainingverifierssolvemath} & \textbf{0.3920} (0.1368, 0.6222) & \textbf{0.0865} (0.0835, 0.0989) & \textbf{0.1931} (10) & \textbf{0.2199} \\
    141 & BBH \protect\cite{suzgun-etal-2023-challenging} & \textbf{0.3226} (0.0587, 0.6181) & \textbf{0.0978} (0.0963, 0.1038) & \textbf{0.2315} (11) & \textbf{0.2135} \\
    142 & HellaSwag \protect\cite{zellers-etal-2019-hellaswag} & \textbf{0.4395} (0.0815, 0.8333) & \textbf{0.0835} (0.0821, 0.0892) & \textbf{0.1263} (4) & \textbf{0.2132} \\
    143 & C-Eval \protect\cite{huang2023ceval} & \textbf{0.2674} (0.0433, 0.5256) & \textbf{0.0935} (0.0923, 0.0983) & \textbf{0.2816} (13) & \textbf{0.2098} \\
    144 & Global PIQA \protect\cite{chang2025globalpiqaevaluatingphysical} & \textbf{0.2926} (0.0335, 0.6000) & \textbf{0.0860} (0.0845, 0.0920) & \textbf{0.2603} (5) & \textbf{0.2085} \\
    145 & CMMLU \protect\cite{li-etal-2024-cmmlu} & \textbf{0.2688} (0.0525, 0.5111) & \textbf{0.1064} (0.1050, 0.1123) & \textbf{0.2339} (10) & \textbf{0.1997} \\
    146 & AIME 2026 \protect\cite{balunovic2025matharena} & \textbf{0.2667} (0.0384, 0.5333) & \textbf{0.0556} (0.0568, 0.0505) & \textbf{0.2873} (6) & \textbf{0.1978} \\
    147 & CRUX-I \protect\cite{pmlr-v235-gu24c} & \textbf{0.1620} (0.0220, 0.3333) & \textbf{0.2978} (0.2985, 0.2950) & \textbf{0.1141} (3) & \textbf{0.1954} \\
    148 & VLMsAreBlind \protect\cite{rahmanzadehgervi2024vision} & \textbf{0.3436} (0.0577, 0.6666) & \textbf{0.0499} (0.0520, 0.0414) & \textbf{0.1849} (3) & \textbf{0.1884} \\
    149 & LOFT(hard retrieval) $\leq 128\text{K}$ \protect\cite{rahmanzadehgervi2024vision} & \textbf{0.2485} (0.0349, 0.5000) & \textbf{0.1545} (0.1537, 0.1578) & \textbf{0.1434} (4) & \textbf{0.1817} \\
    150 & MMLU-Global \protect\cite{singh-etal-2025-globalmmlu} & \textbf{0.2563} (0.0438, 0.5000) & \textbf{0.1187} (0.1192, 0.1166) & \textbf{0.1674} (4) & \textbf{0.1790} \\
    151 & OmniDocBench1.5 (Lower-Better)(0-1) \protect\cite{ouyang2025omnidocbench} & \textbf{0.1448} (0.0269, 0.2857) & \textbf{0.1027} (0.1039, 0.0979) & \textbf{0.2990} (7) & \textbf{0.1785} \\
    152 & MMCU \protect\cite{zeng2023measuringmassivemultitaskchinese} & \textbf{0.3303} (0.0424, 0.6666) & \textbf{0.0705} (0.0708, 0.0692) & \textbf{0.1368} (3) & \textbf{0.1762} \\
    153 & ChartQA \protect\cite{masry-etal-2022-chartqa} & \textbf{0.2102} (0.0428, 0.4000) & \textbf{0.1279} (0.1292, 0.1226) & \textbf{0.1937} (6) & \textbf{0.1755} \\
    154 & CountBench \protect\cite{paiss2023teaching} & \textbf{0.2151} (0.0485, 0.4000) & \textbf{0.0404} (0.0404, 0.0400) & \textbf{0.2639} (6) & \textbf{0.1681} \\
    155 & ARC-Challenge \protect\cite{leolm2023arc_de} & \textbf{0.3395} (0.0529, 0.6666) & \textbf{0.0430} (0.0417, 0.0482) & \textbf{0.1155} (3) & \textbf{0.1627} \\
    156 & CMATH \protect\cite{wei2023cmathlanguagemodelpass} & \textbf{0.3239} (0.0349, 0.6666) & \textbf{0.0552} (0.0562, 0.0515) & \textbf{0.1154} (3) & \textbf{0.1619} \\
    157 & EgoSchema \protect\cite{10.5555/3666122.3668126} & \textbf{0.0000} (0.0087, 0.0000) & \textbf{0.2533} (0.2509, 0.2630) & \textbf{0.1188} (3) & \textbf{0.1281} \\
    158 & DocVQA \protect\cite{mathew2021docvqa} & \textbf{0.1435} (0.0179, 0.3000) & \textbf{0.0476} (0.0472, 0.0493) & \textbf{0.1752} (5) & \textbf{0.1192} \\
\end{longtable}
}
\newpage

\input{checklist.tex}

\end{document}

%% file: checklist.tex
\section*{NeurIPS Paper Checklist}
 

\begin{enumerate}

\item {\bf Claims}
    \item[] Question: Do the main claims made in the abstract and introduction accurately reflect the paper's contributions and scope?
    \item[] Answer: \answerYes{} 
    \item[] Justification: The main claims made in the abstract and introduction accurately reflect the main paper’s contributions and scope.

\item {\bf Limitations}
    \item[] Question: Does the paper discuss the limitations of the work performed by the authors?
    \item[] Answer: \answerYes{} 
    \item[] Justification: This paper discusses the limitation in section~\ref{app:limitations}

\item {\bf Theory assumptions and proofs}
    \item[] Question: For each theoretical result, does the paper provide the full set of assumptions and a complete (and correct) proof?
    \item[] Answer: \answerNA{} 
    \item[] Justification: The paper does not present formal theoretical results requiring proofs; the metric formulations are defined in Section~\ref{sec:mathodology} and detailed in Appendices~\ref{appendix:sdm_derivation} and~\ref{appendix:critic_derivation}.

    \item {\bf Experimental result reproducibility}
    \item[] Question: Does the paper fully disclose all the information needed to reproduce the main experimental results of the paper to the extent that it affects the main claims and/or conclusions of the paper (regardless of whether the code and data are provided or not)?
    \item[] Answer: \answerYes{} 
    \item[] Justification: The paper provides the information needed to reproduce the main results through the metric definitions in Section~\ref{sec:mathodology}, the data pipeline in Section~\ref{data_pipline} and Appendix~\ref{append:Expanded Data Pipeline}, and the robustness settings in Appendices~\ref{app:threshold_sensitivity},~\ref{app:robustness_noise_dropout}, and~\ref{app:antisaturation_sensitivity}.

\item {\bf Open access to data and code}
    \item[] Question: Does the paper provide open access to the data and code, with sufficient instructions to faithfully reproduce the main experimental results, as described in supplemental material?
    \item[] Answer: \answerYes{} 
    \item[] Justification: The paper provides an anonymous repository for the released code and benchmark metadata, while the reproduction-relevant processing steps are described in Section~3 and Appendix~\ref{append:Expanded Data Pipeline}.

\item {\bf Experimental setting/details}
    \item[] Question: Does the paper specify all the training and test details (e.g., data splits, hyperparameters, how they were chosen, type of optimizer) necessary to understand the results?
    \item[] Answer: \answerYes{} 
    \item[] Justification: The experimental settings are specified through the BHI metric definitions in Section~\ref{sec:mathodology}, the data pipeline in Section~\ref{data_pipline}, and the expanded implementation details in Appendix~\ref{append:Expanded Data Pipeline}.

\item {\bf Experiment statistical significance}
    \item[] Question: Does the paper report error bars suitably and correctly defined or other appropriate information about the statistical significance of the experiments?
    \item[] Answer: \answerYes{} 
    \item[] Justification: The paper reports robustness and sensitivity analyses for the main conclusions in Section~\ref{sec:main_results}, Appendix~\ref{app:threshold_sensitivity}, Appendix~\ref{app:robustness_noise_dropout}, Appendix~\ref{app:antisaturation_sensitivity}, and Appendix~\ref{app:coverage_bias}.

\item {\bf Experiments compute resources}
    \item[] Question: For each experiment, does the paper provide sufficient information on the computer resources (type of compute workers, memory, time of execution) needed to reproduce the experiments?
    \item[] Answer: \answerYes{} 
    \item[] Justification: As explained in Appendix~\ref{app:compute}, the experiments do not involve model training or fine-tuning.
    
\item {\bf Code of ethics}
    \item[] Question: Does the research conducted in the paper conform, in every respect, with the NeurIPS Code of Ethics \url{https://neurips.cc/public/EthicsGuidelines}?
    \item[] Answer: \answerYes{} 
    \item[] Justification: The study analyzes publicly reported benchmark results and releases anonymized research artifacts, and we are not aware of any aspect that deviates from the NeurIPS Code of Ethics.

\item {\bf Broader impacts}
    \item[] Question: Does the paper discuss both potential positive societal impacts and negative societal impacts of the work performed?
    \item[] Answer: \answerYes{} 
    \item[] Justification: This paper discusses the impact in Section~\ref{results_and_validation}
    
\item {\bf Safeguards}
    \item[] Question: Does the paper describe safeguards that have been put in place for responsible release of data or models that have a high risk for misuse (e.g., pre-trained language models, image generators, or scraped datasets)?
    \item[] Answer: \answerNA{} 
    \item[] Justification: The paper does not release pretrained models, generative systems, or high-risk scraped datasets; the released artifacts are code and benchmark-level metadata for reproducible analysis.

\item {\bf Licenses for existing assets}
    \item[] Question: Are the creators or original owners of assets (e.g., code, data, models), used in the paper, properly credited and are the license and terms of use explicitly mentioned and properly respected?
    \item[] Answer: \answerYes{} 
    \item[] Justification: All mentioned model technical reports and benchmarks are properly credited.

\item {\bf New assets}
    \item[] Question: Are new assets introduced in the paper well documented and is the documentation provided alongside the assets?
    \item[] Answer: \answerYes{} 
    \item[] Justification: The paper releases new analysis code and benchmark metadata through the anonymous repository, with the construction and validation procedure documented in Section~\ref{sec:mathodology} and Appendix~\ref{append:Expanded Data Pipeline}.

\item {\bf Crowdsourcing and research with human subjects}
    \item[] Question: For crowdsourcing experiments and research with human subjects, does the paper include the full text of instructions given to participants and screenshots, if applicable, as well as details about compensation (if any)? 
    \item[] Answer: \answerNA{} 
    \item[] Justification: The paper does not involve crowdsourcing, user studies, or research with human subjects.

\item {\bf Institutional review board (IRB) approvals or equivalent for research with human subjects}
    \item[] Question: Does the paper describe potential risks incurred by study participants, whether such risks were disclosed to the subjects, and whether Institutional Review Board (IRB) approvals (or an equivalent approval/review based on the requirements of your country or institution) were obtained?
    \item[] Answer: \answerNA{} 
    \item[] Justification: The paper does not involve human subjects, so IRB approval or equivalent review is not applicable.

\item {\bf Declaration of LLM usage}
    \item[] Question: Does the paper describe the usage of LLMs if it is an important, original, or non-standard component of the core methods in this research? Note that if the LLM is used only for writing, editing, or formatting purposes and does \emph{not} impact the core methodology, scientific rigor, or originality of the research, declaration is not required.
    \item[] Answer: \answerYes{} 
    \item[] Justification: The core methodology does not use LLMs as an important, original, or non-standard component; LLMs are only the objects of benchmark analysis in this study.

\end{enumerate}